 \documentclass[pmlr,twocolumn,10pt]{jmlr} 

\usepackage{booktabs}
\usepackage{siunitx}

\usepackage[switch]{lineno}

\usepackage{caption}
\usepackage{graphicx}
\usepackage{subcaption}
\usepackage{array}
\usepackage{ragged2e}
\usepackage{caption}
\usepackage{multirow}
\usepackage{colortbl}
\usepackage{amsmath}
\usepackage{caption}
\usepackage{xcolor}

\theorembodyfont{\upshape}
\theoremheaderfont{\scshape}
\theorempostheader{:}
\theoremsep{\newline}

\jmlrvolume{297}
\jmlryear{2025}
\jmlrworkshop{Machine Learning for Health (ML4H) 2025} 

 \title[m1]{m1: Unleash the Potential of Test-Time Scaling for Medical Reasoning with Large Language Models}

 \author{%
  \Name{Xiaoke Huang} \Email{xhuan192@ucsc.edu}\\
  \addr UC Santa Cruz \\
  \Name{Juncheng Wu} \Email{jwu418@ucsc.edu}\\
  \addr UC Santa Cruz \\
  \Name{Hui Liu} \Email{huiliulayne@gmail.com}\\
  \addr Amazon Research \\
  \Name{Xianfeng Tang} \Email{tangxianfeng@outlook.com}\\
  \addr Amazon Research \\
  \Name{Yuyin Zhou} \Email{yzhou284@ucsc.edu}\\
  \addr UC Santa Cruz
 }

\begin{document}

\maketitle

\begin{figure*}[t]
  \centering
  \includegraphics[width=\textwidth]{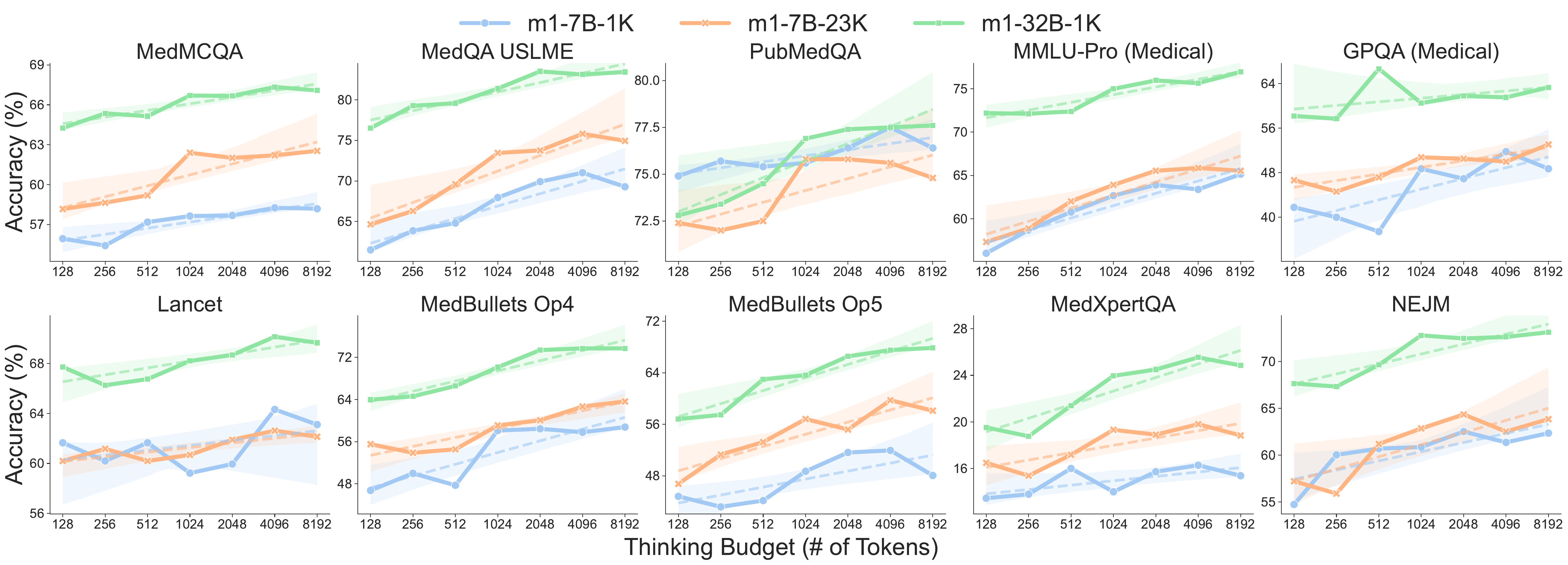}
  \caption{\textbf{Test-time scaling of \texttt{m1} series.} Each plot shows
  accuracy (\%) vs. reasoning token budget for different \texttt{m1} model
  variants on various medical QA datasets. All models improve steadily as the
  thinking length increases, with the 32B model reaching the best accuracy.
  The linear regression lines are dotted with 95\% CIs.}
  \label{fig:intro:teaser}
\end{figure*}

\begin{abstract}
Test-time scaling has emerged as a powerful technique for enhancing the reasoning capabilities of large language models (LLMs). However, its effectiveness in medical reasoning remains uncertain, as the medical domain fundamentally differs from mathematical tasks in terms of knowledge representation and decision-making processes. In this paper, we provide the first comprehensive investigation of test-time scaling for medical reasoning and present \textbf{m1},  a simple yet effective approach that increases a model’s medical reasoning capability at inference. Our evaluation across diverse medical tasks demonstrates that test-time scaling (by increasing the ``thinking'' token budget) consistently enhances medical reasoning, enabling lightweight fine-tuned models under 10B parameters to establish new state-of-the-art performance, while our 32B model achieves results comparable to previous 70B-scale medical LLMs. However, we identify an optimal reasoning token budget of approximately 4K, beyond which performance may degrade due to overthinking. Budget forcing, which extends test-time computation through iterative prompts (e.g., appending ``Wait"), helps models double-check answers but does not necessarily improve the overall medical QA performance and, in some cases, even introduces errors into previously correct responses. Taken together, our case-by-case analysis further identifies insufficient medical knowledge as a key bottleneck that prevents further performance gains through test-time scaling. To overcome this constraint, we find that increasing data scale, improving data quality, and expanding model capacity consistently enhance medical knowledge grounding, enabling continued performance improvements—particularly on challenging medical benchmarks where smaller models reach saturation. These findings underscore fundamental differences between medical and mathematical reasoning in LLMs, highlighting that enriched medical knowledge, other than increased reasoning depth alone, is essential for fully realizing the benefits of test-time scaling.
\end{abstract}


\paragraph*{Data and Code Availability}
Our code, models, and data are publicly available at~\url{https://github.com/UCSC-VLAA/m1}.

\paragraph*{Institutional Review Board (IRB)}
Our research does not require IRB approval.

\section{Introduction}

Test-time scaling has emerged as a promising direction to enhance LLM reasoning by enabling models to ``think more'' during inference~\cite{yang2025towards}. 
OpenAI’s o1~\cite{jaech2024openai} demonstrated that significantly extending an LLM's chain-of-thought can yield remarkable gains in problem-solving ability in both STEM fields and the medical domain~\cite{muennighoff2025s1,xie2024preliminary},
but the exact methodology was not disclosed, spurring many replication efforts.
Among the most successful replication attempts is the open-source s1 method~\cite{muennighoff2025s1}, which achieved remarkable results through a surprisingly simple approach. By fine-tuning a 32B parameter model on just 1K carefully curated examples with reasoning traces and implementing an inference control mechanism via a "Wait" token, {s1} enabled the model to effectively double-check its work. This simple approach produced state-of-the-art results on challenging mathematical benchmarks, \textit{e.g.}, outperformed OpenAI's o1-preview by up to 27\%.

Despite these advances, applying test-time scaling to the medical domain remains largely underexplored~\cite{jiang2025meds}.
The medical domain presents unique challenges for LLMs: questions often involve multi-step logical reasoning, accurate recall of medical knowledge, and careful consideration to avoid unsafe or harmful answers~\cite{chen2024huatuogpt}. As the medical field fundamentally differs from mathematical tasks in terms of knowledge representation and decision-making processes, the effectiveness of test-time scaling for medical reasoning remains uncertain.

While advanced proprietary models like GPT-4~\cite{hurst2024gpt} and Med-PaLM~\cite{singhal2022large} have achieved expert-level scores on medical exams~\cite{zhang2024ultramedical}, open-source medical LLMs still struggle to reliably solve complex medical problems. Improving reasoning in these models is critical, as healthcare applications demand not just factual accuracy but robust diagnostic and therapeutic reasoning capabilities.
Existing medical reasoning LLMs such as HuatuoGPT-o1~\cite{chen2024huatuogpt} typically rely on computationally intensive methods like reinforcement learning with verification mechanisms. 
This raises a key question: \textit{Can a simple test-time scaling strategy, with minimal fine-tuning, also unlock strong medical reasoning?}

In this paper, we answer in the affirmative by presenting \texttt{m1}, a lightweight methodology that adapts the test-time scaling paradigm to medical QA tasks. Our approach is straightforward: we curate a high-quality set of medical questions with detailed step-by-step solutions (only 1K / 23K examples), fine-tune open LLMs on this data, and at inference use test-time controls to ensure the model fully ``thinks through'' problems before answering. Figure~\ref{fig:intro:teaser} illustrates the outcome: as we allow the model to generate longer chains of thought (x-axis increasing), accuracy on various medical benchmarks consistently improves for our \texttt{m1} models. Notably, even our 7B-parameter model fine-tuned on 1K examples shows significant gains with more reasoning steps, and our 32B model achieves the highest scores across the board.

To better understand the impact of test-time scaling on medical reasoning in LLMs, we conduct a fine-grained study, systematically examining the effects of \textbf{thinking budgets, inference techniques, data curation}, and \textbf{model capacity}. While increasing the token budget consistently improves performance, we identify an optimal reasoning threshold of approximately 4K tokens, beyond which accuracy declines due to overthinking. In addition to increasing the token budget (Figure~\ref{fig:intro:teaser}), reasoning can also be extended through budget forcing, wherein the model iteratively prolongs its thought process during inference~\cite{muennighoff2025s1}. However, unlike in mathematical reasoning—where iterative refinement often enhances accuracy—forcing additional reasoning in medical QA yields limited benefits and, in some cases, even degrades performance. This occurs when models with erroneous knowledge reconsider correct responses during extended reasoning, ultimately arriving at incorrect conclusions.

A closer analysis of failure cases reveals that this bottleneck stems from deficiencies in essential medical knowledge, which cannot be resolved merely by increasing the thinking budget. Consequently, extending the reasoning window reaches a fundamental limit, and budget-forcing techniques offer negligible benefits, as models lacking foundational knowledge remain anchored to incorrect assumptions. Even with additional reasoning steps, these models still struggle to retrieve accurate information.
In such cases, improving data quality and increasing model capacity provide more effective avenues for improvement. 
Our thorough ablation of data filtering strategies, dataset size, and model scaling demonstrates that when scaling thinking budget reaches its bottleneck, further performance gains can be achieved by enhancing data quality and scaling the model. Specifically, larger, difficulty-filtered, and diversity-sampled datasets consistently improve performance, while larger models further enhance scalability. This is because larger-capacity models or those fine-tuned on more extensive, high-quality datasets inherently possess richer medical knowledge, leading to higher accuracy. In conclusion, test-time scaling alone is insufficient for enhancing medical reasoning in LLMs—it needs to be complemented by scaling model size and improving knowledge grounding through high-quality data.

Our 7B model fine-tuned on 23K examples (\texttt{m1-7B-23K}) attains new state-of-the-art accuracy of 60.32\% among in-domain and out-domain medical exam datasets, surpassing previously established specialized models of similar scale such as HuatuoGPT-o1-7B/8B (trained with complex RL on 40K instances)~\cite{chen2024huatuogpt} and UltraMedical-8B (trained on hundreds of thousands of medical instructions)~\cite{zhang2024ultramedical}. 
Furthermore, our larger 32B model trained with only 1K fine-tuning samples (\texttt{m1-32B-1K}) achieves performance comparable to 2X bigger resource-exhausted models (around 70B parameters with high training costs), underscoring the efficiency of our test-time scaling approach.
All data, code, and models are publicly available to encourage future exploration in optimizing inference strategies in clinical AI applications.
\section{Related Works}

\paragraph{Test-time scaling for LLMs.} There is a growing interest in techniques that enhance an LLM's reasoning without altering its weights, by allocating more computation at inference time~\cite{jaech2024openai,meng2023deepscaler,OpenReasonerZero2025}. A basic form is chain-of-thought prompting, e.g. instructing the model to “think step by step,” which often improves performance on complex tasks~\cite{wei2022chain}. More explicit approaches include generating multiple solutions and using majority voting or self-consistency to pick an answer, or employing search-based strategies with verifiers and lookahead. These methods trade extra inference passes for accuracy gains. In contrast, sequential test-time scaling keeps a single reasoning thread but makes it longer. OpenAI's o1 model hinted at the power of simply extending the reasoning length~\cite{jaech2024openai}. \cite{muennighoff2025s1} formalized this by fine-tuning an LLM to utilize special “Wait” tokens, which allow controlling response length during inference. Their budget forcing method (described below) proved more effective than parallel voting strategies. Other recent research has proposed optimizing the allocation of test-time compute, for example finding an optimal stopping length per problem to avoid overthinking~\cite{yang2025towards}. Our work builds directly on the simple test-time scaling idea~\cite{muennighoff2025s1,aggarwal2025l1} by extending thinking traces with ``wait'' --- we apply it to a new domain (medicine) and confirm its benefits in a very different setting. We focus on single-trace sequential reasoning, noting that it is complementary to orthogonal advances like tool use or retrieval augmentation.

\paragraph{Medical LLMs.} The success of GPT-4 in medical exams~\cite{zhang2024ultramedical} has spurred numerous open efforts to train medical domain LLMs~\cite{lucas2024reasoning,xiong2024improving}. Early approaches centered on domain-specific pre-training: e.g. \cite{wu2024pmc} and \cite{qiu2024towards} pre-trained Llama models on medical text corpora (MIMIC-III~\cite{johnson2016mimic}, PubMed~\footnote{\url{https://pubmed.ncbi.nlm.nih.gov/}}, etc.) to inject medical knowledge. While this improves knowledge recall, the gains on reasoning-heavy tasks were limited~\cite{jiang2025meds,medreason}. More recent projects emphasize instruction tuning and reinforcement learning specialized for medicine. For example, OpenBioLLM was fine-tuned with expert-validated instructions and Direct Preference Optimization, and reportedly outperformed GPT-4 and Med-PaLM-2 on several biomedical QA benchmarks~\cite{pal2024openbiollms}. Med42 is another open model suite that achieved impressive results, even exceeding GPT-4.0 on many multi-choice medical QA tasks~\cite{christophe2024med42}. To push reasoning ability further, some works incorporate explicit reasoning supervision or verification. HuatuoGPT-o1 introduced verifiable medical problem-solving: they constructed 40K problems with known solutions and used a two-stage training (SFT + RL with a verifier) to train a 70B model~\cite{chen2024huatuogpt}. This model achieved new state-of-the-art results on medical reasoning benchmarks, outperforming both general and prior medical LLMs~\cite{zhang2024ultramedical}. UltraMedical built a massive dataset of ~410K mixed manual/synthetic instructions for biomedicine~\cite{jiang2025meds}, and fine-tuned Llama-3 models with supervised and preference learning. The 70B UltraMedical model reached 86.5\% accuracy on MedQA, nearly matching Med-PaLM2~\cite{singhal2025toward} and GPT-4~\cite{achiam2023gpt}. 
In contrast, our approach remains lightweight as we do not introduce new RL or verification components, and our dataset size (1K-23K) is relatively small, yet through test-time scaling, we achieve competitive results with these state-of-the-art models. We hope this encourages more exploration of inference-time techniques as an efficient alternative for domain-specific LLMs.

\section{Method}

\begin{figure}[t]
\begin{center}
\makebox[\linewidth][c]{
\includegraphics[width=1.0\linewidth]{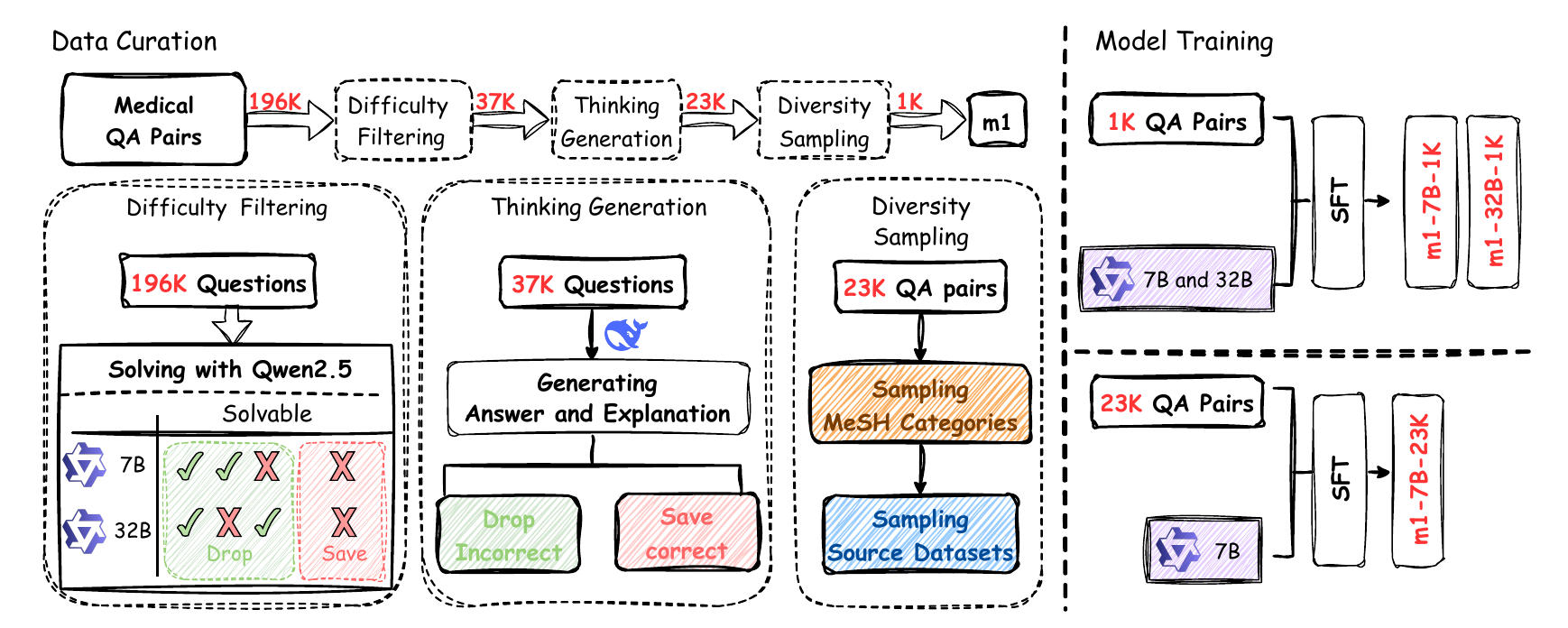}
}
\end{center}
\caption{
\textbf{An overview of our data curation and training pipeline.} We start with 196K raw medical QA examples, apply difficulty filtering (retaining 37K that Qwen2.5-7B-Instruct~\cite{yang2024qwen2} or its 32B version cannot solve), then use DeepSeek-R1~\cite{guo2025deepseek} to generate reasoning and keep correct solutions (\texttt{m23K}). We perform diversity sampling to select a 1K high-quality subset (\texttt{m1K}). These datasets are used to fine-tune base models (Qwen2.5 7B and 32B Instruct) via Supervised Fine-Tuning (SFT), resulting in the m1 models (\texttt{m1-7B-1K}, \texttt{m1-7B-23K}, \texttt{m1-32B-1K}).
}
\label{fig:method:method}
\end{figure}


\subsection{Data Curation}

\label{sec3:method:data}

\paragraph{Initial collection.} To construct the training data for \texttt{m1} through a multi-step refinement process, we begin with a large pool of approximately 196K medical QA samples compiled from public datasets: MedMCQA~\cite{pal2022medmcqa}, MedQA-USMLE~\cite{jin2021disease}, HeadQA~\cite{vilares2019head}, and PubMedQA~\cite{jin2019pubmedqa}. These include multiple-choice questions from medical exams as well as open-ended research questions. 
All samples are decontaminated against the evaluation data in Section~\ref{sec4:exp:eval:datasets}. 
More details are presented in Appendix~\ref{app:data_stats}. 

\paragraph{Difficulty filtering.} 
Following s1~\cite{muennighoff2025s1}, we identify a subset of solvable yet non-trivial problems by performing difficulty filtering using two strong base models.
Specifically, we use Qwen2.5-Instruct~\cite{yang2024qwen2} (an open general LLM) of 7B and 32B parameters to attempt each question. We filter a question if either Qwen-7B or Qwen-32B answers it correctly. This heuristic retains questions that are challenging to solve, eliminating those that are too easy for either models. Difficulty filtering pruned the dataset from 196K down to ~37K samples.

\paragraph{Thinking generation.} We employ DeepSeek-R1~\cite{guo2025deepseek}, a state-of-the-art open reasoning LLM, to generate a chain-of-thought and final answer for each of the 37K questions. DeepSeek-R1 was chosen for its robust reasoning capability (it's comparable to OpenAI's o1~\cite{jaech2024openai} in multi-step problem solving). For each question, we prompt DeepSeek-R1 to produce a detailed solution explanation ending in a definitive answer. We then apply solution validation: we only keep those instances where DeepSeek-R1's final answer is correct (matching the ground-truth). This yields a set of ~23K high-quality ``thinking - answers'', where each question now paired with a verified-correct reasoning process.
This step ensures our training data predominantly consists of valid reasoning, while incorrect chains are discarded. More details are presented in Appendix~\ref{app:data_stats} and~\ref{app:impl_details:data_gen}.

\paragraph{Diversity sampling.} 
We design a diversity sampling strategy to construct a well-balanced and enriched subset for training our primary model. This process highlights two key components: domain balance and dataset balance. First, we ensure domain balance by annotating each sample with Medical Subject Headings (MeSH) categories\footnote{\url{https://www.ncbi.nlm.nih.gov/mesh/}} (See Appendix for the details), enabling systematic coverage across medical specialties (e.g., cardiology, neurology) and question types. Second, we address dataset imbalance through stratified sampling, first selecting domains, then source datasets, and finally individual samples (see Appendix Tables~\ref{app:tab:data_stats} and~\ref{app:tab:domain_stats} for distributions).
We perform stratified sampling at the dataset level to address the imbalance in sample counts across datasets (Appendix, Table~\ref{app:tab:data_stats},~\ref{app:tab:domain_stats}).
Specifically, we first sample a domain, then sample a dataset, and finally roll-out a sample.
The process is repeated until there are 1K samples (\texttt{m1K}), which will be served as our core training set for \texttt{m1}. 
The remaining 23K difficult samples (\texttt{m23K}) can be used to augment training or for ablations. 
We provide summary statistics of the final data in Appendix~\ref{app:data_stats}.

\subsection{Model Training}
\label{sec3:method:train}
We fine-tune three model variants, corresponding to two model sizes (7B and 32B) and two training set sizes (1K and 23K). For each, we use the pre-trained Qwen2.5-Instruct model as the initialization. Qwen2.5 is a recent high-performance open LLM~\cite{yang2024qwen2}; using it as our base ensures strong general language ability and allows us to focus on injecting medical reasoning. We format each training example in a ``question $\to$ reasoning $\to$ answer” style. 
This format teaches the model to produce a coherent reasoning process and then give the answer. 
Using this data, we perform SFT for each model:

$\bullet$ \texttt{m1-7B-1K}: Fine-tuned on the 1K \texttt{m1K} dataset using the Qwen2.5-7B-Instruct. This represents the minimal training scenario.

$\bullet$  \texttt{m1-7B-23K}: Fine-tuned on the full 23K filtered dataset using Qwen2.5-7B-Instruct. This lets us examine the effect of more training data (23K vs 1K) at the same model size.

$\bullet$   \texttt{m1-32B-1K}: Fine-tuned on the 1K dataset using the larger Qwen2.5-32B-Instruct. This shows the effect of a larger model with minimal data.

\subsection{Inference}
\label{sec3:method:inference}
At inference time, we employ test-time scaling by managing the model's generation of the chain-of-thought. Specifically, we define a thinking budget: a maximum number of tokens the model is allowed to generate before producing a final answer. By allocating a larger budget, we give the model more ``thinking space'' to potentially reason through the problem. If the model would naturally finish its reasoning early, we intervene to use the budget fully. We also apply budget forcing technique to extend the thinking process of the model: when the model outputs \verb|end-of-think| token indicating the end of thinking before reaching the token budget, we replace it with ``Wait.'' and force the model to keep the generation of the thinking traces. According to~\cite{muennighoff2025s1}, this method often leads the model to double-check or refine its initial answers for math problems.

\section{Experiments}
\begin{table*}[t]
    \centering
    \resizebox{\textwidth}{!}{%
    \begin{tabular}{>{\scriptsize}l|ccc|ccccccc|c}
    \toprule
    \multicolumn{1}{l|}{Model}                     & {MedMC} & {MedQA} & {PubMed} & {MMLU-P} & {GPQA} & {Lancet} & {MedB (4)} & {MedB (5)} & {MedX} & {NEJM} & {Avg.} \\ \cline{2-12}
                              & \multicolumn{3}{c|}{In-Distribution Test}                  & \multicolumn{7}{c|}{Out-of-Distribution Test}                                                                                          &         \\
    \midrule
    \multicolumn{12}{c}{$<$  10B LLMs} \\
    \midrule
    MedLlama3-8B-v1          & {\cellcolor[rgb]{0.988,0.988,1}}34.74                                 & {\cellcolor[rgb]{0.988,0.988,1}}55.07                                & {\cellcolor[rgb]{0.988,0.988,1}}52.70                            & {\cellcolor[rgb]{0.988,0.988,1}}27.43                                     & {\cellcolor[rgb]{0.992,0.992,1}}30.77                                 & {\cellcolor[rgb]{0.988,0.988,1}}42.23                   & {\cellcolor[rgb]{0.988,0.988,1}}38.31                             & {\cellcolor[rgb]{0.988,0.988,1}}33.77                             & {\cellcolor[rgb]{0.988,0.988,1}}11.04                        & {\cellcolor[rgb]{0.988,0.988,1}}49.25                 & {\cellcolor[rgb]{0.988,0.988,1}}37.53                      \\
    MedLlama3-8B-v2          & {\cellcolor[rgb]{0.871,0.949,0.89}}59.34                              & {\cellcolor[rgb]{0.996,0.996,1}}59.39                                & 75.50                                                            & {\cellcolor[rgb]{0.996,0.996,1}}55.11                                     & {\cellcolor[rgb]{0.992,0.992,1}}36.41                                 & {\cellcolor[rgb]{0.992,0.992,1}}52.43                   & {\cellcolor[rgb]{0.996,0.996,1}}48.38                             & {\cellcolor[rgb]{0.992,0.992,1}}39.94                             & {\cellcolor[rgb]{0.992,0.992,1}}13.46                        & {\cellcolor[rgb]{0.992,0.992,1}}54.56                 & {\cellcolor[rgb]{0.996,0.996,1}}49.45                      \\
    OpenBioLLM-8B            & {\cellcolor[rgb]{0.996,0.996,1}}54.63                                 & {\cellcolor[rgb]{0.988,0.988,1}}55.30                                & {\cellcolor[rgb]{0.996,0.996,1}}70.10                            & {\cellcolor[rgb]{0.992,0.992,1}}49.32                                     & {\cellcolor[rgb]{0.996,0.996,1}}41.03                                 & {\cellcolor[rgb]{0.992,0.992,1}}52.43                   & {\cellcolor[rgb]{0.988,0.988,1}}41.23                             & {\cellcolor[rgb]{0.988,0.988,1}}32.47                             & {\cellcolor[rgb]{0.996,0.996,1}}14.29                        & {\cellcolor[rgb]{0.992,0.992,1}}54.23                 & {\cellcolor[rgb]{0.992,0.992,1}}46.50                      \\
    MMed-8B                  & {\cellcolor[rgb]{0.996,0.996,1}}52.71                                 & {\cellcolor[rgb]{0.988,0.988,1}}54.28                                & {\cellcolor[rgb]{0.992,0.992,1}}63.40                            & {\cellcolor[rgb]{0.992,0.992,1}}48.27                                     & {\cellcolor[rgb]{0.992,0.992,1}}34.87                                 & {\cellcolor[rgb]{0.992,0.992,1}}53.40                   & {\cellcolor[rgb]{0.988,0.988,1}}41.23                             & {\cellcolor[rgb]{0.988,0.988,1}}35.39                             & {\cellcolor[rgb]{0.992,0.992,1}}13.73                        & {\cellcolor[rgb]{0.992,0.992,1}}54.39                 & {\cellcolor[rgb]{0.992,0.992,1}}45.17                      \\
    MMedS-8B                 & {\cellcolor[rgb]{0.992,0.992,1}}47.29                                 & {\cellcolor[rgb]{0.992,0.992,1}}57.19                                & {\cellcolor[rgb]{0.737,0.89,0.776}}77.50                         & {\cellcolor[rgb]{0.988,0.988,1}}33.55                                     & {\cellcolor[rgb]{0.988,0.988,1}}22.05                                 & {\cellcolor[rgb]{0.996,0.996,1}}55.10                   & {\cellcolor[rgb]{0.933,0.973,0.945}}54.22                         & {\cellcolor[rgb]{0.651,0.855,0.706}}55.84                         & {\cellcolor[rgb]{0.706,0.878,0.749}}17.39                    & {\cellcolor[rgb]{0.988,0.988,1}}53.40                 & {\cellcolor[rgb]{0.992,0.992,1}}47.35                      \\
    MMed-8B-EnIns            & 58.09                                                                 & {\cellcolor[rgb]{0.996,0.996,1}}60.33                                & {\cellcolor[rgb]{0.992,0.992,1}}63.80                            & {\cellcolor[rgb]{0.996,0.996,1}}51.60                                     & {\cellcolor[rgb]{0.996,0.996,1}}45.90                                 & {\cellcolor[rgb]{0.996,0.996,1}}55.34                   & {\cellcolor[rgb]{0.678,0.867,0.729}}59.09                         & {\cellcolor[rgb]{0.631,0.847,0.686}}56.17                         & {\cellcolor[rgb]{0.553,0.816,0.624}}18.56                    & 62.35                                                 & 53.12                                                      \\
    Med42-8B                 & {\cellcolor[rgb]{0.996,0.996,1}}56.35                                 & {\cellcolor[rgb]{0.996,0.996,1}}59.78                                & {\cellcolor[rgb]{0.937,0.973,0.945}}76.00                        & {\cellcolor[rgb]{0.996,0.996,1}}55.64                                     & {\cellcolor[rgb]{0.98,0.992,0.984}}48.21                              & {\cellcolor[rgb]{0.996,0.996,1}}59.47                   & {\cellcolor[rgb]{0.992,0.992,1}}44.81                             & {\cellcolor[rgb]{0.996,0.996,1}}46.75                             & {\cellcolor[rgb]{0.996,0.996,1}}14.63                        & {\cellcolor[rgb]{0.957,0.984,0.961}}62.69             & {\cellcolor[rgb]{0.996,0.996,1}}52.43                      \\
    UltraMedical-8B-3        & {\cellcolor[rgb]{0.886,0.953,0.902}}59.22                             & {\cellcolor[rgb]{0.592,0.831,0.655}}71.09                            & {\cellcolor[rgb]{0.996,0.996,1}}71.10                            & {\cellcolor[rgb]{0.969,0.988,0.973}}61.50                                 & {\cellcolor[rgb]{0.831,0.933,0.859}}50.00                             & {\cellcolor[rgb]{0.89,0.953,0.906}}61.89                & {\cellcolor[rgb]{0.933,0.973,0.945}}54.22                         & {\cellcolor[rgb]{0.894,0.957,0.91}}52.27                          & {\cellcolor[rgb]{0.984,0.996,0.988}}15.25                    & {\cellcolor[rgb]{0.706,0.878,0.753}}64.51             & {\cellcolor[rgb]{0.749,0.898,0.788}}56.11                  \\
    UltraMedical-8B-3.1      & {\cellcolor[rgb]{0.412,0.757,0.502}}63.78                             & {\cellcolor[rgb]{0.392,0.749,0.486}}75.73                            & {\cellcolor[rgb]{0.51,0.796,0.584}}79.20                         & {\cellcolor[rgb]{0.686,0.871,0.733}}64.30                                 & {\cellcolor[rgb]{0.937,0.976,0.949}}48.72                             & {\cellcolor[rgb]{0.388,0.745,0.482}}67.23               & {\cellcolor[rgb]{0.388,0.745,0.482}}64.61                         & {\cellcolor[rgb]{0.698,0.875,0.745}}55.19                         & {\cellcolor[rgb]{0.706,0.878,0.749}}17.39                    & {\cellcolor[rgb]{0.388,0.745,0.482}}66.83             & {\cellcolor[rgb]{0.392,0.749,0.486}}60.30                  \\
    HuatuoGPT-o1-7B          & {\cellcolor[rgb]{0.443,0.769,0.529}}63.47                             & {\cellcolor[rgb]{0.573,0.824,0.639}}71.56                            & {\cellcolor[rgb]{0.588,0.831,0.655}}78.60                        & {\cellcolor[rgb]{0.388,0.745,0.482}}67.23                                 & 47.95                                                                 & {\cellcolor[rgb]{0.867,0.945,0.886}}62.14               & 52.92                                                             & 50.65                                                             & 15.11                                                        & {\cellcolor[rgb]{0.616,0.843,0.675}}65.17             & {\cellcolor[rgb]{0.631,0.847,0.69}}57.48                   \\
    HuatuoGPT-o1-8B          & {\cellcolor[rgb]{0.388,0.745,0.482}}63.97                             & {\cellcolor[rgb]{0.435,0.765,0.522}}74.78                            & {\cellcolor[rgb]{0.388,0.745,0.482}}80.10                        & {\cellcolor[rgb]{0.745,0.894,0.784}}63.71                                 & {\cellcolor[rgb]{0.388,0.745,0.482}}55.38                             & {\cellcolor[rgb]{0.663,0.859,0.714}}64.32               & {\cellcolor[rgb]{0.714,0.882,0.757}}58.44                         & {\cellcolor[rgb]{0.914,0.965,0.929}}51.95                         & {\cellcolor[rgb]{0.776,0.91,0.812}}16.84                     & {\cellcolor[rgb]{0.663,0.859,0.714}}64.84             & {\cellcolor[rgb]{0.467,0.78,0.549}}59.43                   \\
    \hline
    Qwen2.5-7B-Instruct      & {\cellcolor[rgb]{0.996,0.996,1}}56.56                                 & 61.51                                                                & {\cellcolor[rgb]{0.996,0.996,1}}71.30                            & 61.17                                                                     & {\cellcolor[rgb]{0.996,0.996,1}}42.56                                 & {\cellcolor[rgb]{0.957,0.984,0.965}}61.17               & {\cellcolor[rgb]{0.992,0.992,1}}46.75                             & {\cellcolor[rgb]{0.992,0.992,1}}40.58                             & {\cellcolor[rgb]{0.988,0.988,1}}12.15                        & {\cellcolor[rgb]{0.996,0.996,1}}59.04                 & {\cellcolor[rgb]{0.996,0.996,1}}51.28                      \\
    \quad+CoT  & {\cellcolor[rgb]{0.996,0.996,1}}56.11                                 & {\cellcolor[rgb]{0.875,0.949,0.894}}64.49                            & {\cellcolor[rgb]{0.996,0.996,1}}72.60                            & {\cellcolor[rgb]{0.902,0.961,0.918}}62.15                                 & {\cellcolor[rgb]{0.624,0.843,0.682}}52.56                             & 60.68                                                   & {\cellcolor[rgb]{0.996,0.996,1}}50.97                             & {\cellcolor[rgb]{0.992,0.992,1}}42.86                             & {\cellcolor[rgb]{0.992,0.992,1}}13.18                        & {\cellcolor[rgb]{0.996,0.996,1}}58.54                 & {\cellcolor[rgb]{0.976,0.992,0.98}}53.41                   \\
    \textbf{m1-7B-1K}                 & {\cellcolor[rgb]{0.984,0.996,0.988}}58.26                             & {\cellcolor[rgb]{0.596,0.831,0.659}}71.01                            & {\cellcolor[rgb]{0.737,0.89,0.776}}77.50                         & {\cellcolor[rgb]{0.6,0.835,0.663}}65.15                                   & {\cellcolor[rgb]{0.686,0.871,0.733}}51.79                             & {\cellcolor[rgb]{0.663,0.859,0.714}}64.32               & {\cellcolor[rgb]{0.694,0.875,0.741}}58.77                         & {\cellcolor[rgb]{0.914,0.965,0.929}}51.95                         & {\cellcolor[rgb]{0.847,0.937,0.871}}16.29                    & {\cellcolor[rgb]{0.98,0.992,0.98}}62.52               & {\cellcolor[rgb]{0.608,0.839,0.671}}57.76                  \\
    \textbf{m1-7B-23K}                & {\cellcolor[rgb]{0.537,0.808,0.612}}62.54                             & {\cellcolor[rgb]{0.388,0.745,0.482}}75.81                            & {\cellcolor[rgb]{0.961,0.984,0.969}}75.80                        & {\cellcolor[rgb]{0.529,0.804,0.6}}65.86                                   & {\cellcolor[rgb]{0.58,0.827,0.643}}53.08                              & {\cellcolor[rgb]{0.82,0.925,0.847}}62.62                & {\cellcolor[rgb]{0.439,0.769,0.525}}63.64                         & {\cellcolor[rgb]{0.388,0.745,0.482}}59.74                         & {\cellcolor[rgb]{0.388,0.745,0.482}}19.81                    & {\cellcolor[rgb]{0.729,0.89,0.773}}64.34              & {\cellcolor[rgb]{0.388,0.745,0.482}}60.32                  \\
    \bottomrule
    \multicolumn{12}{c}{$>$  10B LLMs} \\
    \toprule
    Qwen2.5-72B-Instruct       & {\cellcolor[rgb]{0.996,0.996,1}}66.60                                 & {\cellcolor[rgb]{0.996,0.996,1}}74.55                                & {\cellcolor[rgb]{0.988,0.988,1}}70.80                            & {\cellcolor[rgb]{0.992,0.992,1}}66.06                                     & {\cellcolor[rgb]{0.996,0.996,1}}62.05                                 & {\cellcolor[rgb]{0.996,0.996,1}}66.50                   & {\cellcolor[rgb]{0.996,0.996,1}}57.14                             & {\cellcolor[rgb]{0.996,0.996,1}}53.57                             & {\cellcolor[rgb]{0.988,0.988,1}}14.91                        & {\cellcolor[rgb]{0.996,0.996,1}}68.99                 & {\cellcolor[rgb]{0.996,0.996,1}}60.12                      \\
    \quad+CoT   & {\cellcolor[rgb]{0.996,0.996,1}}66.15                                 & {\cellcolor[rgb]{0.973,0.992,0.98}}76.43                             & {\cellcolor[rgb]{0.992,0.992,1}}71.30                            & {\cellcolor[rgb]{0.996,0.996,1}}69.77                                     & 63.85                                                                 & {\cellcolor[rgb]{0.996,0.996,1}}65.78                   & {\cellcolor[rgb]{0.996,0.996,1}}60.06                             & {\cellcolor[rgb]{0.996,0.996,1}}54.22                             & {\cellcolor[rgb]{0.988,0.988,1}}14.84                        & {\cellcolor[rgb]{0.996,1,0.996}}69.15                 & {\cellcolor[rgb]{0.996,0.996,1}}61.16                      \\
    Med42-70B                & {\cellcolor[rgb]{0.988,0.988,1}}62.28                                 & {\cellcolor[rgb]{0.988,0.988,1}}51.14                                & {\cellcolor[rgb]{0.961,0.984,0.965}}78.10                        & {\cellcolor[rgb]{0.988,0.988,1}}54.53                                     & {\cellcolor[rgb]{0.988,0.988,1}}50.77                                 & {\cellcolor[rgb]{0.988,0.988,1}}54.61                   & {\cellcolor[rgb]{0.988,0.988,1}}45.78                             & {\cellcolor[rgb]{0.988,0.988,1}}37.99                             & {\cellcolor[rgb]{0.992,0.992,1}}16.29                        & {\cellcolor[rgb]{0.988,0.988,1}}56.05                 & {\cellcolor[rgb]{0.988,0.988,1}}50.75                      \\
    OpenBioLLM-70B           & {\cellcolor[rgb]{0.549,0.812,0.62}}74.23                              & {\cellcolor[rgb]{0.996,0.996,1}}75.10                                & {\cellcolor[rgb]{0.753,0.898,0.792}}79.30                        & {\cellcolor[rgb]{0.996,0.996,1}}71.92                                     & {\cellcolor[rgb]{0.988,0.988,1}}50.77                                 & {\cellcolor[rgb]{0.914,0.965,0.929}}68.93               & {\cellcolor[rgb]{0.996,0.996,1}}58.44                             & {\cellcolor[rgb]{0.996,1,0.996}}54.55                             & {\cellcolor[rgb]{0.796,0.918,0.827}}21.33                    & {\cellcolor[rgb]{0.996,0.996,1}}67.83                 & {\cellcolor[rgb]{0.969,0.988,0.973}}62.24                  \\
    UltraMedical-70B-3       & {\cellcolor[rgb]{0.627,0.847,0.686}}72.94                             & {\cellcolor[rgb]{0.624,0.843,0.682}}83.90                            & {\cellcolor[rgb]{0.631,0.847,0.69}}80.00                         & {\cellcolor[rgb]{0.996,0.996,1}}73.94                                     & {\cellcolor[rgb]{0.992,0.992,1}}58.72                                 & {\cellcolor[rgb]{0.388,0.745,0.482}}75.49               & {\cellcolor[rgb]{0.576,0.824,0.643}}72.08                         & {\cellcolor[rgb]{0.667,0.863,0.718}}64.61                         & {\cellcolor[rgb]{0.769,0.906,0.804}}21.67                    & {\cellcolor[rgb]{0.659,0.859,0.71}}73.13              & {\cellcolor[rgb]{0.616,0.843,0.675}}67.65                  \\
    HuatuoGPT-o1-70B         & {\cellcolor[rgb]{0.486,0.788,0.565}}75.23                             & {\cellcolor[rgb]{0.486,0.788,0.565}}86.80                            & {\cellcolor[rgb]{0.388,0.745,0.482}}81.40                        & {\cellcolor[rgb]{0.827,0.929,0.855}}76.09                                 & {\cellcolor[rgb]{0.388,0.745,0.482}}66.67                             & {\cellcolor[rgb]{0.604,0.835,0.667}}72.82               & {\cellcolor[rgb]{0.576,0.824,0.643}}72.08                         & {\cellcolor[rgb]{0.537,0.808,0.612}}68.51                         & {\cellcolor[rgb]{0.388,0.745,0.482}}26.36                    & {\cellcolor[rgb]{0.573,0.824,0.639}}74.13             & {\cellcolor[rgb]{0.463,0.776,0.545}}70.01                  \\
    HuatuoGPT-o1-72B         & {\cellcolor[rgb]{0.388,0.745,0.482}}76.76                             & {\cellcolor[rgb]{0.388,0.745,0.482}}88.85                            & {\cellcolor[rgb]{0.647,0.855,0.702}}79.90                        & {\cellcolor[rgb]{0.388,0.745,0.482}}80.46                                 & {\cellcolor[rgb]{0.89,0.957,0.91}}64.36                               & {\cellcolor[rgb]{0.761,0.902,0.796}}70.87               & {\cellcolor[rgb]{0.388,0.745,0.482}}77.27                         & {\cellcolor[rgb]{0.388,0.745,0.482}}73.05                         & {\cellcolor[rgb]{0.62,0.843,0.678}}23.53                     & {\cellcolor[rgb]{0.388,0.745,0.482}}76.29             & {\cellcolor[rgb]{0.388,0.745,0.482}}71.13                  \\
    \midrule
    Qwen2.5-32B-Instruct     & {\cellcolor[rgb]{0.992,0.992,1}}64.83                                 & {\cellcolor[rgb]{0.996,0.996,1}}75.26                                & {\cellcolor[rgb]{0.988,0.988,1}}68.00                            & {\cellcolor[rgb]{0.965,0.984,0.969}}74.72                                 & 63.85                                                                 & {\cellcolor[rgb]{0.996,0.996,1}}66.02                   & 60.39                                                             & {\cellcolor[rgb]{0.996,0.996,1}}52.92                             & {\cellcolor[rgb]{0.988,0.988,1}}13.87                        & {\cellcolor[rgb]{0.996,0.996,1}}66.67                 & {\cellcolor[rgb]{0.996,0.996,1}}60.65                      \\
    \quad+CoT & {\cellcolor[rgb]{0.992,0.992,1}}64.33                                 & {\cellcolor[rgb]{0.996,0.996,1}}74.86                                & {\cellcolor[rgb]{0.988,0.988,1}}68.90                            & {\cellcolor[rgb]{0.965,0.984,0.969}}74.72                                 & {\cellcolor[rgb]{0.78,0.91,0.816}}64.87                               & {\cellcolor[rgb]{0.996,0.996,1}}66.75                   & 60.39                                                             & {\cellcolor[rgb]{0.996,0.996,1}}54.22                             & {\cellcolor[rgb]{0.988,0.988,1}}14.56                        & {\cellcolor[rgb]{0.996,0.996,1}}66.33                 & {\cellcolor[rgb]{0.996,0.996,1}}60.99                      \\
    \textbf{m1-32B-1K}                & {\cellcolor[rgb]{0.98,0.992,0.984}}67.34                              & {\cellcolor[rgb]{0.643,0.851,0.698}}83.50                            & {\cellcolor[rgb]{0.996,0.996,1}}77.60                            & {\cellcolor[rgb]{0.741,0.894,0.78}}76.94                                  & {\cellcolor[rgb]{0.388,0.745,0.482}}66.67                             & {\cellcolor[rgb]{0.816,0.925,0.847}}70.15               & {\cellcolor[rgb]{0.518,0.8,0.592}}73.70                           & {\cellcolor[rgb]{0.561,0.82,0.627}}67.86                          & {\cellcolor[rgb]{0.459,0.776,0.541}}25.53                    & {\cellcolor[rgb]{0.659,0.859,0.71}}73.13              & {\cellcolor[rgb]{0.576,0.824,0.643}}68.24                 \\
    \bottomrule
    \end{tabular}
    }
    \caption{\textbf{Baseline Comparisons.} We report accuracy (\%) on each evaluation dataset for various models. Our m1 models (in bold) are shown in the $\leq10B$ group (m1-7B variants) and $>10B$ group (m1-32B). ``+CoT'' indicates using chain-of-thought prompting at inference for that base model. We mark Green color within each parameter group: the deeper the color, the higher the accuracy. For header abbreviations, please refer to Section~\ref{sec4:exp:eval:datasets}. }
    \label{tab:model_comparison}
\end{table*}

\subsection{Evaluation Settings}

\label{sec4:exp:eval:datasets}
\paragraph{Datasets.}

We evaluate on nine medical QA benchmarks, grouped into In-Distribution and Out-of-Distribution tests. We measure accuracy for all datasets.
1) In-Distribution tests: The accompanying test splits from our training data, include MedMCQA~\cite{pal2022medmcqa} (MedMC); MedQA-USMLE~\cite{jin2021disease} (MedQA), and PubMedQA~\cite{jin2019pubmedqa}  (PubMed).
2) Out-of-Distribution tests: The datasets are not included in training and stylistically distinct, assessing \texttt{m1}'s reasoning generalization: medical related questions from MMLU-Pro~\cite{wang2024mmlu} (MMLU-P) and GPQA~\cite{rein2024gpqa}, small QA sets from Lancet and the New England Journal of Medicine (NEJM); 4 Options (MEdB (4)) and 5 Options (MedB (5)) splits from the MedBullets platform~\cite{chen2024benchmarking}; and MedXpertQA~\cite{zuo2025medxpertqa} (MedX). 

\paragraph{LLM baselines.}

We compare our models against a variety of general and specialized medical LLM baselines: 1) general base instruct models Qwen2.5-7B, tested both as-is and with chain-of-thought prompting (+CoT); 2) specialized medical models including MedLlama3~\footnote{\url{https://huggingface.co/johnsnowlabs/}}, OpenBioLLM~\cite{pal2024openbiollms}, MMed-Llama~\cite{qiu2024towards}, Med42~\cite{christophe2024med42}, UltraMedical~\cite{zhang2024ultramedical}; 3) medical reasoning model HuatuoGPT-o1~\cite{chen2024huatuogpt}, which undergoes complex RL training. 

Additionally, we compare to state-of-the-art large open medical LLMs (\textgreater 10B), including Med42-70B~\cite{christophe2024med42}, OpenBioLLM-70B~\cite{pal2024openbiollms}, UltraMedical-70B~\cite{zhang2024ultramedical}, HuatuoGPT-o1-70B/72B~\cite{chen2024huatuogpt}, and baseline Qwen2.5 models (32B, 72B) with their respective +CoT versions. HuatuoGPT-o1 and UltraMedical employ complex training strategies involving reinforcement learning or expert feedback; therefore, matching or exceeding their performance with our simpler test-time scaling method underscores its effectiveness.

\subsection{Results}

\paragraph{Test-time scaling with different thinking budgets.} We first evaluate how increasing the chain-of-thought token budget at inference affects performance on various medical QA datasets. As illustrated by the upward trajectories in Figure~\ref{fig:intro:teaser}, our m1 approach gains consistent accuracy improvements as the thinking budget grows, demonstrating the efficacy of simple test-time scaling. 
Despite simplicity, Table~\ref{tab:model_comparison} presents that our \texttt{m1-7B-23K} achieves an average accuracy of 60.32\% amongst in-distribution and out-of-distribution sets, which exceeds complex RL tuned HuatuoGPT-o1-7B by 2.84\%, and matches the large-scale SFT-tuned UltraMedical-8B.
Notably, beyond 4K tokens, the improvements begin to saturate, indicating limited additional benefit from extremely long reasoning. 

\paragraph{Larger model capacity helps.} When scaling model size from 7B to 32B parameters, we observe a more pronounced benefit from test-time scaling as larger models inherently possess richer medical knowledge.
In Table~\ref{tab:model_comparison}, \texttt{m1-32B-1K} consistently outperforms or matches even larger (70B+) specialized medical LLMs, demonstrating that pairing a larger base model with simple supervised thinking traces and inference-time scaling yields strong results. 
This trend is also apparent in Figure~\ref{fig:intro:teaser}, where the 32B model’s accuracy curve leads across most datasets as the thinking budget increases.

\begin{figure}[t]
\begin{center}
\includegraphics[width=\linewidth]{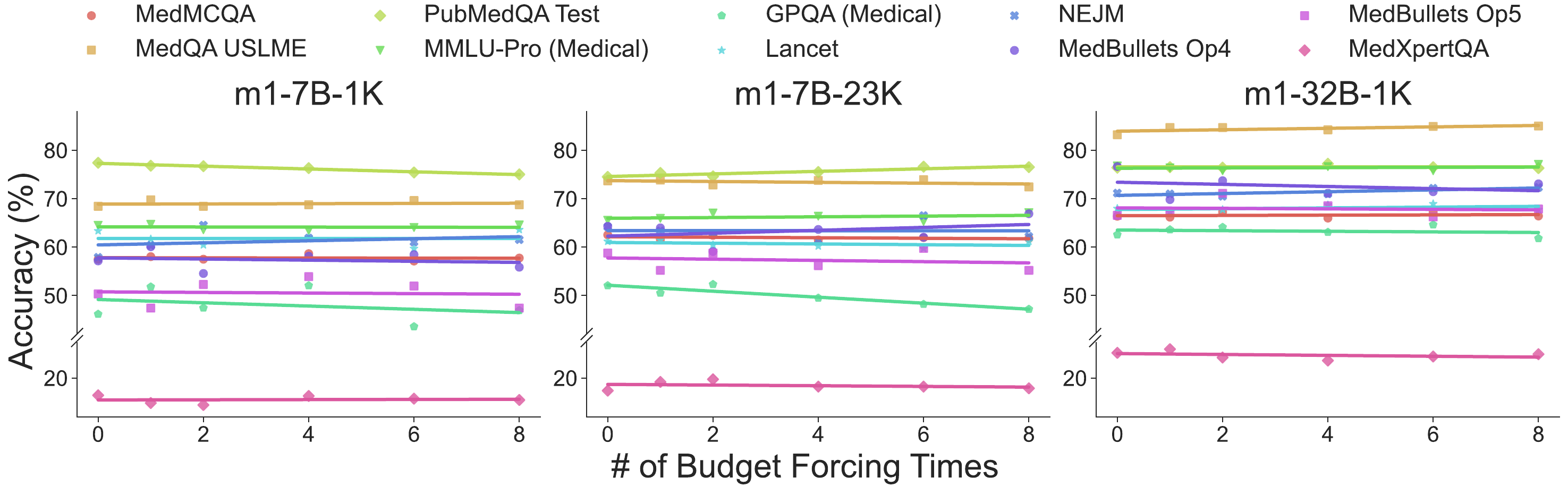}
\end{center}
\caption{
\textbf{Force thinking for different evaluation datasets.} Accuracy vs. number of budget forcing times (iterations of injecting ``Wait'') for each m1 model (\texttt{7B-1K}, \texttt{7B-23K}, \texttt{32B-1K}). A value of 0 means the model’s first answer is taken without forcing, while higher values mean the model was compelled to reconsider up to that many times (within a 2048-token limit).
}
\label{fig:keep_thinking}
\end{figure}

\paragraph{Budget forcing does not help.} Unlike mathematical tasks, where prompting the model to repeatedly refine its chain-of-thought can yield further gains, our experiments show diminishing returns from forced re-thinking (Figure~\ref{fig:keep_thinking}). Although the model will generate additional intermediate tokens when repeatedly prompted to ``keep thinking'', we see minimal improvement, suggesting that medical reasoning may differ from math domains in how additional iterative reasoning is best leveraged. We analyze such failure cases in the following sections.

\paragraph{SFT data ablation.}
We ablate two key data curation steps used in our SFT process: difficulty filtering and diversity sampling (comprising domain and dataset balancing). As shown in Table~\ref{tab:data_ablation}, at the 1K training scale, models fine-tuned on difficulty-filtered data outperform those trained on randomly sampled data by +0.27\% percentage points on average. Adding domain balance further improves performance by 0.43\%, and incorporating both domain and dataset balance yields additional gains, reaching up to 56.55\% average accuracy. At the 23K scale, overall performance improves substantially, and difficulty filtering alone provides a +0.65\% accuracy boost on average. These results highlight the critical role of both data quality and scale in enhancing model performance.

\begin{table*}[t]
    \centering
    \resizebox{\textwidth}{!}{%
    \begin{tabular}{>{\scriptsize}l|ccc|ccccccc|c}
    \toprule
    \multicolumn{1}{l|}{Data Filtering}                     & {MedMC} & {MedQA} & {PubMed} & {MMLU-P} & {GPQA} & {Lancet} & {MedB (4)} & {MedB (5)} & {MedX} & {NEJM} & {Avg.} \\ \cline{2-12}
                              & \multicolumn{3}{c|}{In-Distribution Test}                  & \multicolumn{7}{c|}{Out-of-Distribution Test}                                                                                          &         \\
    \midrule
    \multicolumn{12}{c}{1K Scale} \\
    \midrule
Random                            & 57.26             & 66.46            & 73.60        & 61.63                 & 44.62             & 62.62 & 54.55         & 53.57         & 14.70    & 58.54 & 54.75 \\
Hard Random                       & 56.99             & 68.66            & 73.70        & 63.26                 & 46.41             & 62.14 & 56.82         & 44.81         & 16.36    & 61.03 & 55.02 \\
Hard Domain                & \textbf{58.88}             & 66.38            & 74.00        & \textbf{64.95}                 & 45.38             & 63.11 & 54.55         & 49.68         & 16.36    & \textbf{61.19} & 55.45 \\
Hard Domain Dataset & 57.97             & \textbf{70.23}            & \textbf{76.10}        & 64.23                 & \textbf{49.74}             & \textbf{62.14} & \textbf{57.79}         & \textbf{50.97}         & \textbf{17.12}    & 59.20 & \textbf{56.55} \\
\midrule
    \multicolumn{12}{c}{23K Scale} \\
\midrule
Random                           & 60.41             & 71.64            & \textbf{76.50}        & \textbf{67.43}                 & 48.72             & \textbf{62.38} & 60.06         & 54.22         & 15.32    & 61.86 & 57.85 \\
Hard                      & \textbf{62.01 }            & \textbf{73.76}            & 75.80        & 65.54                 & \textbf{50.51}             & 61.89 & 60.06         & \textbf{55.19}         & \textbf{18.91}    & \textbf{64.34} & \textbf{58.80} \\
    \bottomrule
    \end{tabular}
    }
    \caption{
    \textbf{Data Filtering Ablation: difficulty filtering (``Hard''), and ``Domain'' and ``Dataset'' balance for diversity sampling.} 
    Difficulty filtering consistently yields the largest gains across both 1K and 23K training scales, while domain and dataset balancing provide complementary improvements. Notably, scaling up from 1K to 23K substantially boosts accuracy, underscoring the importance of data scale.
    The same header abbreviations as Table~\ref{tab:model_comparison}.
    }
    \label{tab:data_ablation}
\end{table*}

\begin{figure}[t]
\begin{center}
\includegraphics[width=\linewidth]{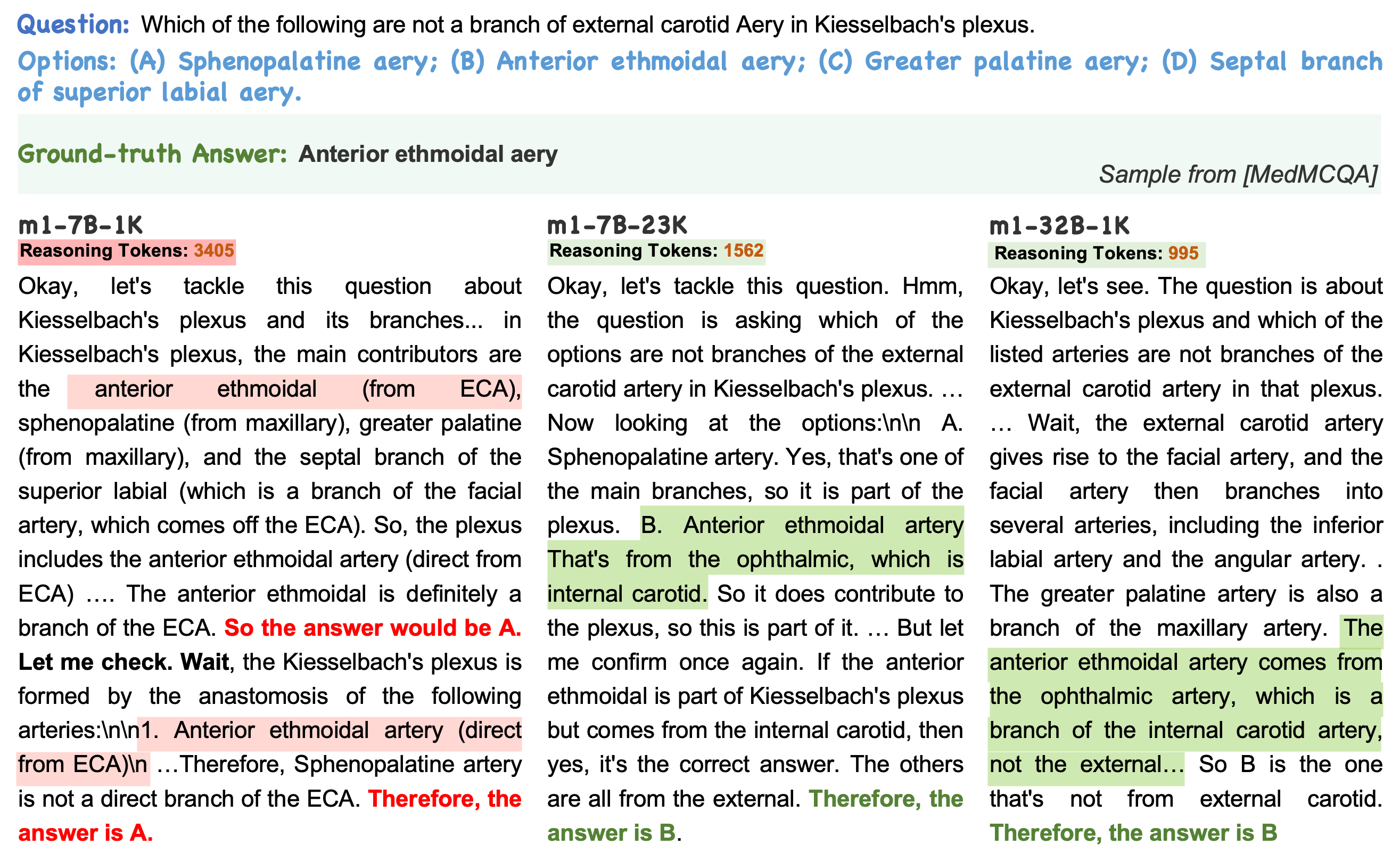}
\end{center}
\caption{
\textbf{A failure case of test-time scaling} with the Qwen2.5-7B using 1K reasoning data. Although the m1-7B-1K conducts the longest reasoning, its deficiency in essential medical knowledge prevents it from producing the right answer. On the other hand, both m1-7B-23K and m1-32B-1K effectively resolve the question with a relatively brief reasoning procedure.
}
\label{fig:case_knowledge}
\end{figure}

\paragraph{Failure cases.}
In this section, we discuss several fail cases where the models fail to perform test-time scaling, underscoring the \textit{critical role of accurate knowledge in medical reasoning models}.
Specifically, our investigation can be distilled into the following key points:

$\bullet$ \textbf{The extent of knowledge is crucial for effective medical reasoning.} As illustrated in Figure~\ref{fig:case_knowledge}, m1-7B-1K is unable to generate accurate reasoning because it lacks crucial knowledge: 'Anterior ethmoidal artery belongs to the internal carotid'. In contrast, both m1-7B-23K and m1-32B-1K possess this essential knowledge. Consequently, having a substantial amount of accurate knowledge, whether from fine-tuning data or the pre-training model, enhances the model's capability for medical reasoning. This is similarly evidenced in Table~\ref{tab:model_comparison},  where m1-7B-23K and m1-32B-1K exhibit a significantly better performance.

\begin{figure}[t]
\begin{center}
\includegraphics[width=\linewidth]{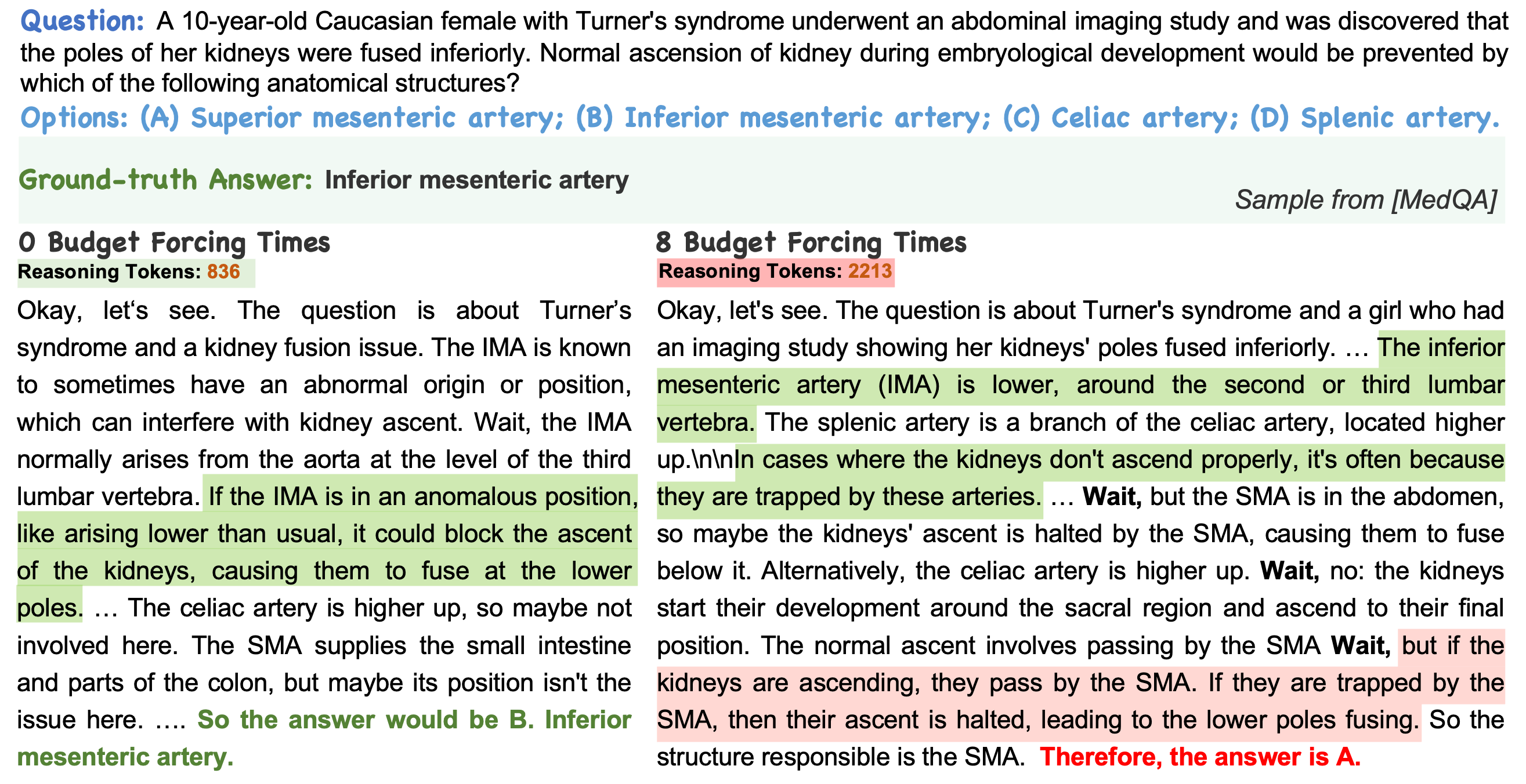}
\end{center}
\caption{
\textbf{A failure case of budget forcing.} Initially, the model produces the correct answer, but forcing it to re-think causes the model to retrieve incorrect knowledge, ultimately resulting in an erroneous answer.}
\label{fig:case_budget_times_fail_success_first}
\end{figure}

$\bullet$ \textbf{Incorrect knowledge obstructs the reasoning.} As demonstrated in Figure~\ref{fig:case_budget_times_fail_success_first}, even when the model generates the correct answer at first, forcing it to re-think cause it to retrieve faulty information, which results in an incorrect response. Therefore, such erroneous knowledge may lead to unstable reasoning process, highlighting the importance of verifying the accuracy of the training data for medical reasoning models.

$\bullet$ \textbf{Test-time scaling fails to rectify incorrect knowledge.} In fields like math and coding, scaling up thinking processes can enhance a model's reasoning by allowing it to conduct self-reflection and identify errors in its previous logic. However, in the medical domain, errors largely stem from misconceptions in knowledge. These are difficult to correct merely by increasing the reasoning budget. As illustrated in Figure~\ref{fig:case_knowledge}, despite m1-7B-1K executing the most extended reasoning, its lack of crucial medical knowledge hinders it from arriving at the correct answer. Additionally, in Appendix Figure~\ref{fig:case_budget_fail_always}, even when the model is forced to re-think multiple times, it is unable to rectify the inaccurate knowledge.

\section{Conclusions}

We introduced \texttt{m1}, demonstrating that test-time scaling significantly improves medical reasoning in large language models without requiring extensive fine-tuning. Performance across diverse medical QA benchmarks consistently improved with increased inference-time reasoning budgets. Crucially, we found that test-time scaling alone cannot remedy fundamental deficiencies in medical knowledge, emphasizing the necessity for high-quality medical data and model scale expansion.  
m1 achieves strong performance, outperforming more expensive approaches such as HuatuoGPT-o1 and UltraMedical on various benchmarks. Our 7B model trained on 23K data establishes a new state-of-the-art in the $\leq10B$ parameter category, and our 32B model rivals models $2\times$ in size.
We release a full-stack open-source package including the curated dataset (m1K), fine-tuned model weights, and inference code with budget control, to encourage future exploration in
optimizing inference strategies in clinical AI applications.


\bibliography{main}

\appendix

\clearpage

\section{Data Statistics}

\label{app:data_stats}

\paragraph{Dataset statistics.} Table~\ref{app:tab:data_stats} shows the statistics of all datasets used in the paper. Note that the sample counts across datasets are highly imbalanced.

\begin{table*}[h]
\centering
\caption{The statistics of all datasets used in the paper.}
\label{app:tab:data_stats}
\resizebox{\textwidth}{!}{%
\begin{tabular}{l|cccc|c}
\toprule
Dataset                                 & MedQA  & HeadQA & MedMCQA & PubMedQA & Summation \\
\midrule
\rowcolor{gray!10}
Initial collection                      & 10,178 & 2,657  & 182,822 & 500      & 196,157   \\
+After difficulty filtering              & 2,099  & 331    & 35,270  & 116      & 37,816    \\
\rowcolor{gray!10}
+Generating thinking data                & 1,628  & 209    & 21,628  & 39       & 23,504    \\
\midrule
+Decontamination \& deduplication (m23K) & 1,628  & 209    & 21,628  & 28       & 23,493    \\
\rowcolor{gray!10}
Random 23K                              & 1,316  & 317    & 21,831  & 29       & 23,493    \\
\midrule
Random 1K                               & 61     & 8      & 929     & 2        & 1,000     \\
\rowcolor{gray!10}
Hard random 1K                          & 78     & 10     & 909     & 3        & 1,000     \\
Hard Domain Balanced 1K                 & 52     & 20     & 924     & 4        & 1,000     \\
\rowcolor{gray!10}
Hard Domain Dataset Balanced 1K (m1K)   & 274    & 123    & 575     & 28       & 1,000   \\
\bottomrule
\end{tabular}%
}
\end{table*}

\paragraph{Token length statistics.}

\begin{figure}[h]
  \centering

  \subfigure[\texttt{m1K}\label{app:fig:hist_token_m1}]{
    \includegraphics[width=0.495\textwidth]{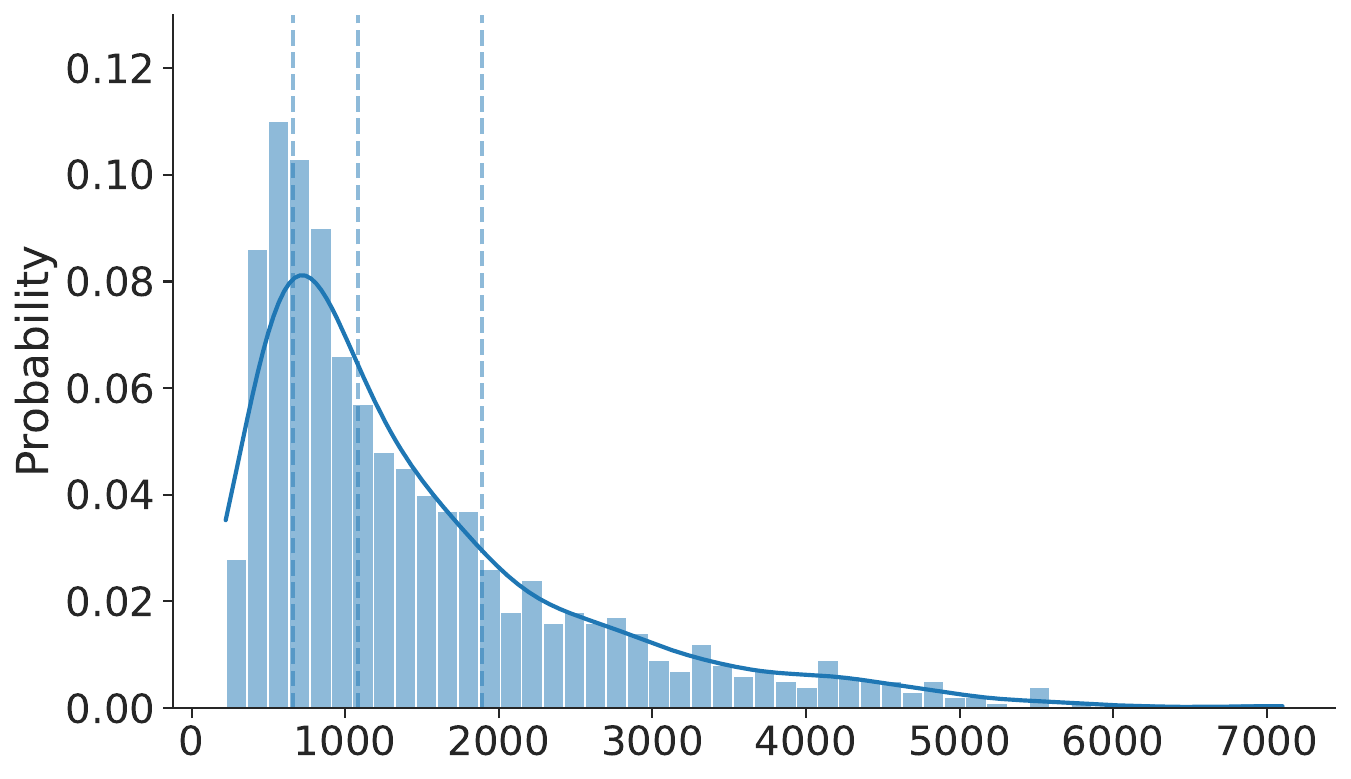}
  }
  \hfill
  \subfigure[\texttt{m23K}\label{app:fig:hist_token_m23k}]{
    \includegraphics[width=0.495\textwidth]{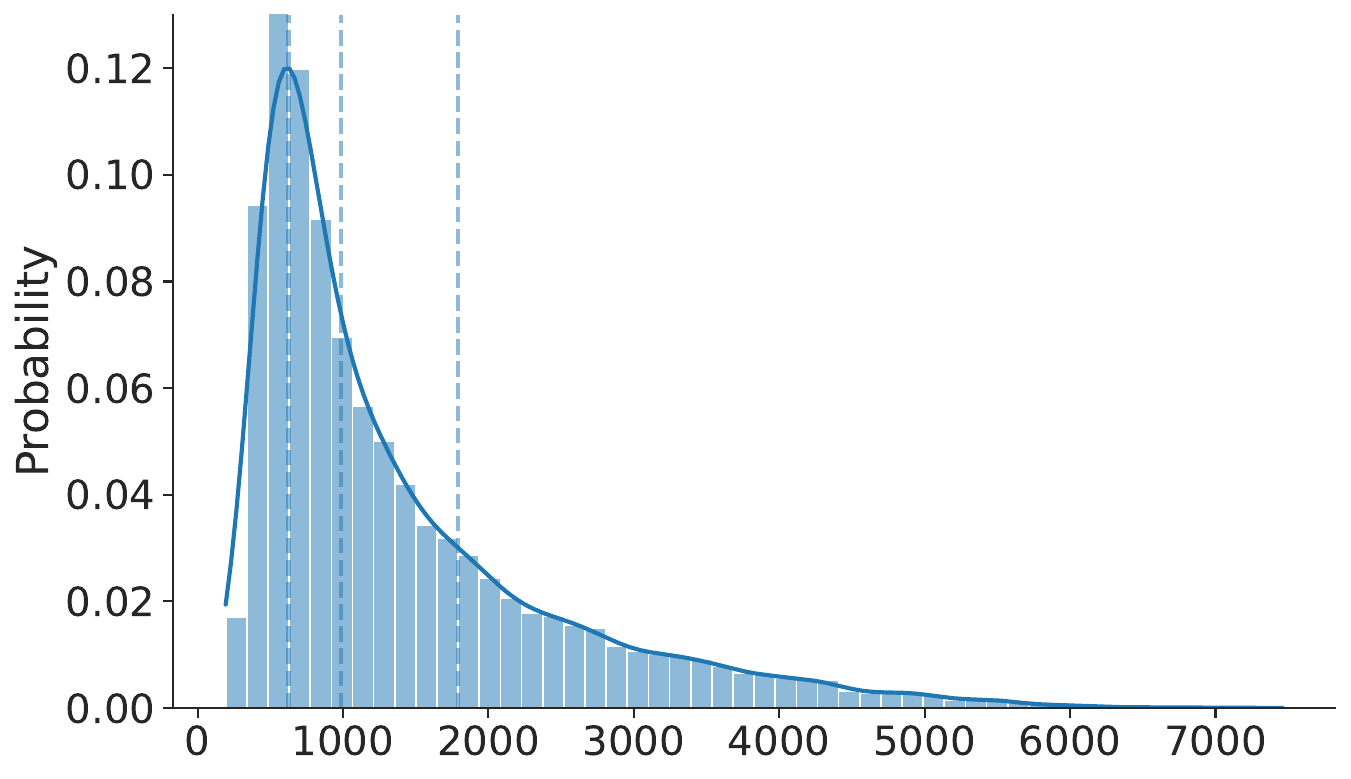}
  }


  \subfigure[s1K\label{app:fig:hist_token_s1}]{
    \includegraphics[width=0.495\textwidth]{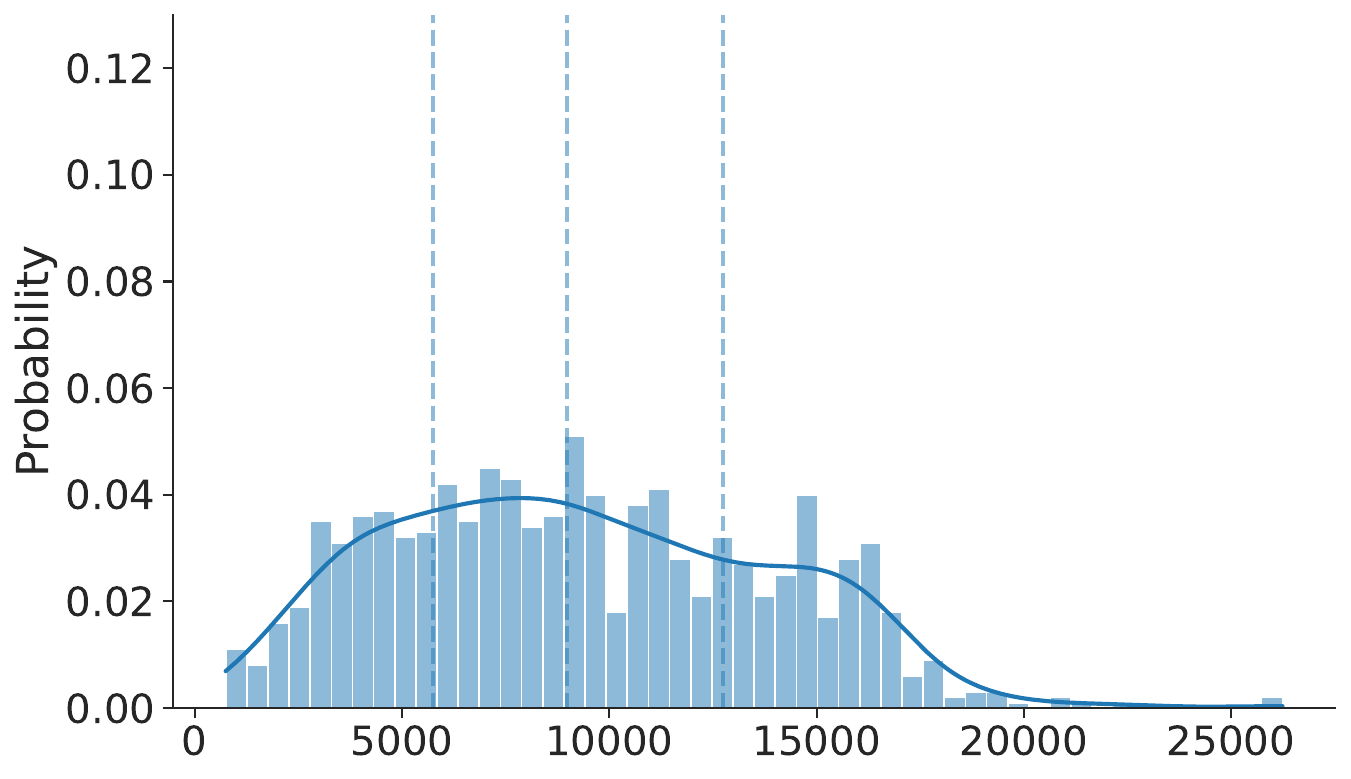}
  }

  \caption{The token length distributions of \textit{m1K} and s1K~\cite{muennighoff2025s1}.
  The 25\%/50\%/75\% quantile is marked in transparent vertical dotted lines.}
  \label{app:fig:hist_token}
\end{figure}

As illustrated in Figure~\ref{app:fig:hist_token}, the distributions of training token lengths between the m1K/m23K and s1K~\cite{muennighoff2025s1} datasets exhibit clear differences. The m1K/m23K dataset shows a strongly right-skewed distribution, with most samples having token lengths clustered around 1,000 tokens, quickly diminishing toward lengths beyond 3,000 tokens. In contrast, the s1K dataset displays a more uniform and broader distribution, spanning widely from about 2,500 to over 15,000 tokens, with peaks around 5,000 to 10,000 tokens. These contrasting distributions reflect differing data preparation strategies: m1K/m23K focuses on concise medical knowledge without longer thinking steps, whereas s1K includes longer, more detailed reasoning traces suitable for complex multi-step inference tasks.

\paragraph{Sample domain statistics.} We list the statistics of sample domains for \texttt{m1K} and \texttt{m23K} in Table~\ref{app:tab:domain_stats}. We use the domain label from 
MeSH Qualifiers with Scope Notes~\footnote{\url{https://www.nlm.nih.gov/mesh/qualifiers_scopenotes.html}}.

\begin{table*}[th]
\centering
\caption{The statistics of sample domain for \texttt{m1K} and \texttt{m23k}.}
\label{app:tab:domain_stats}
\makebox[\linewidth][c]{
\resizebox{0.95\textwidth}{!}{%
\begin{tabular}{l|ccccc|ccccc}
\toprule
\multirow{2}{*}{Domain}        & \multicolumn{5}{c|}{m23K}                        & \multicolumn{5}{c}{m1K}                         \\
 \cline{2-11} 
                               & MedQA & PubMedQA & HeadQA & MedMCQA & \multicolumn{1}{|c|}{Summation} & MedQA & PubMedQA & HeadQA & MedMCQA & \multicolumn{1}{|c}{Summation} \\
\midrule
\rowcolor{gray!10}
Abnormalities                  & 24    & 1        & 0      & 646     & 671       & 10    & 1        & 0      & 10      & 21        \\
Administration \& Dosage       & 9     & 0        & 3      & 590     & 602       & 6     & 0        & 3      & 9       & 18        \\
\rowcolor{gray!10}
Adverse Effects                & 116   & 2        & 2      & 936     & 1,056     & 9     & 2        & 2      & 7       & 20        \\
Agonists                       & 2     & 0        & 1      & 20      & 23        & 2     & 0        & 1      & 13      & 16        \\
\rowcolor{gray!10}
Analogs \& Derivatives         & 1     & 0        & 0      & 10      & 11        & 1     & 0        & 0      & 10      & 11        \\
Analysis                       & 2     & 2        & 14     & 222     & 240       & 2     & 2        & 2      & 6       & 12        \\
\rowcolor{gray!10}
Anatomy \& Histology           & 17    & 0        & 3      & 2,328   & 2,348     & 10    & 0        & 3      & 11      & 24        \\
Antagonists \& Inhibitors      & 14    & 0        & 2      & 113     & 129       & 7     & 0        & 2      & 6       & 15        \\
\rowcolor{gray!10}
Biosynthesis                   & 4     & 0        & 3      & 73      & 80        & 4     & 0        & 3      & 6       & 13        \\
Blood                          & 45    & 1        & 0      & 360     & 406       & 7     & 1        & 0      & 12      & 20        \\
\rowcolor{gray!10}
Blood Supply                   & 19    & 0        & 0      & 228     & 247       & 6     & 0        & 0      & 8       & 14        \\
Cerebrospinal Fluid            & 7     & 0        & 0      & 34      & 41        & 4     & 0        & 0      & 10      & 14        \\
\rowcolor{gray!10}
Chemical Synthesis             & 0     & 0        & 5      & 1       & 6         & 0     & 0        & 5      & 1       & 6         \\
Chemically Induced             & 16    & 0        & 1      & 82      & 99        & 5     & 0        & 1      & 8       & 14        \\
\rowcolor{gray!10}
Chemistry                      & 0     & 0        & 21     & 342     & 363       & 0     & 0        & 7      & 8       & 15        \\
Classification                 & 0     & 0        & 5      & 341     & 346       & 0     & 0        & 5      & 8       & 13        \\
\rowcolor{gray!10}
Complications                  & 73    & 1        & 2      & 625     & 701       & 6     & 1        & 2      & 3       & 12        \\
Congenital                     & 59    & 0        & 0      & 382     & 441       & 7     & 0        & 0      & 4       & 11        \\
\rowcolor{gray!10}
Cytology                       & 3     & 0        & 1      & 90      & 94        & 3     & 0        & 1      & 18      & 22        \\
Deficiency                     & 58    & 1        & 2      & 245     & 306       & 6     & 1        & 2      & 3       & 12        \\
\rowcolor{gray!10}
Diagnosis                      & 229   & 1        & 17     & 1,681   & 1,928     & 2     & 1        & 3      & 4       & 10        \\
Diagnostic Imaging             & 25    & 4        & 2      & 720     & 751       & 7     & 4        & 2      & 11      & 24        \\
\rowcolor{gray!10}
Diet Therapy                   & 3     & 1        & 2      & 41      & 47        & 2     & 1        & 2      & 3       & 8         \\
Drug Effects                   & 12    & 0        & 0      & 107     & 119       & 11    & 0        & 0      & 9       & 20        \\
\rowcolor{gray!10}
Drug Therapy                   & 164   & 0        & 6      & 509     & 679       & 1     & 0        & 4      & 5       & 10        \\
Economics                      & 1     & 1        & 1      & 17      & 20        & 1     & 1        & 1      & 11      & 14        \\
\rowcolor{gray!10}
Education                      & 0     & 1        & 1      & 20      & 22        & 0     & 1        & 1      & 12      & 14        \\
Embryology                     & 12    & 0        & 1      & 245     & 258       & 7     & 0        & 1      & 6       & 14        \\
\rowcolor{gray!10}
Enzymology                     & 5     & 0        & 3      & 95      & 103       & 4     & 0        & 1      & 2       & 7         \\
Epidemiology                   & 8     & 0        & 3      & 265     & 276       & 4     & 0        & 3      & 4       & 11        \\
\rowcolor{gray!10}
Ethics                         & 8     & 0        & 2      & 20      & 30        & 5     & 0        & 2      & 2       & 9         \\
Ethnology                      & 0     & 0        & 0      & 5       & 5         & 0     & 0        & 0      & 5       & 5         \\
\rowcolor{gray!10}
Etiology                       & 133   & 1        & 5      & 657     & 796       & 4     & 1        & 5      & 3       & 13        \\
Genetics                       & 66    & 0        & 5      & 293     & 364       & 4     & 0        & 2      & 3       & 9         \\
\rowcolor{gray!10}
Growth \& Development          & 8     & 1        & 3      & 245     & 257       & 0     & 1        & 1      & 5       & 7         \\
History                        & 0     & 0        & 2      & 72      & 74        & 0     & 0        & 2      & 5       & 7         \\
\rowcolor{gray!10}
Immunology                     & 17    & 0        & 2      & 199     & 218       & 10    & 0        & 2      & 7       & 19        \\
Injuries                       & 12    & 0        & 1      & 430     & 443       & 12    & 0        & 1      & 10      & 23        \\
\rowcolor{gray!10}
Innervation                    & 10    & 0        & 1      & 157     & 168       & 3     & 0        & 1      & 9       & 13        \\
Instrumentation                & 0     & 0        & 0      & 151     & 151       & 0     & 0        & 0      & 17      & 17        \\
\rowcolor{gray!10}
Isolation \& Purification      & 0     & 0        & 0      & 7       & 7         & 0     & 0        & 0      & 7       & 7         \\
Legislation \& Jurisprudence   & 0     & 0        & 1      & 204     & 205       & 0     & 0        & 1      & 15      & 16        \\
\rowcolor{gray!10}
Metabolism                     & 15    & 0        & 5      & 158     & 178       & 1     & 0        & 4      & 5       & 10        \\
Methods                        & 1     & 1        & 3      & 260     & 265       & 1     & 1        & 3      & 9       & 14        \\
\rowcolor{gray!10}
Microbiology                   & 27    & 0        & 0      & 358     & 385       & 5     & 0        & 0      & 8       & 13        \\
Mortality                      & 3     & 0        & 0      & 30      & 33        & 3     & 0        & 0      & 5       & 8         \\
\rowcolor{gray!10}
Nursing                        & 0     & 0        & 5      & 6       & 11        & 0     & 0        & 5      & 6       & 11        \\
Organization \& Administration & 0     & 1        & 3      & 111     & 115       & 0     & 1        & 3      & 10      & 14        \\
\rowcolor{gray!10}
Parasitology                   & 5     & 0        & 1      & 209     & 215       & 5     & 0        & 1      & 9       & 15        \\
Pathogenicity                  & 10    & 0        & 0      & 71      & 81        & 5     & 0        & 0      & 4       & 9         \\
\rowcolor{gray!10}
Pathology                      & 68    & 0        & 2      & 1,532   & 1,602     & 11    & 0        & 2      & 7       & 20        \\
Pharmacokinetics               & 4     & 0        & 6      & 99      & 109       & 3     & 0        & 6      & 1       & 10        \\
\rowcolor{gray!10}
Pharmacology                   & 31    & 0        & 1      & 315     & 347       & 5     & 0        & 1      & 6       & 12        \\
Physiology                     & 45    & 1        & 17     & 1,168   & 1,231     & 5     & 1        & 5      & 13      & 24        \\
\rowcolor{gray!10}
Physiopathology                & 92    & 1        & 3      & 1,093   & 1,189     & 5     & 1        & 3      & 6       & 15        \\
Poisoning                      & 17    & 0        & 0      & 171     & 188       & 5     & 0        & 0      & 7       & 12        \\
\rowcolor{gray!10}
Prevention \& Control          & 17    & 0        & 1      & 131     & 149       & 6     & 0        & 1      & 10      & 17        \\
Psychology                     & 20    & 1        & 21     & 165     & 207       & 0     & 1        & 1      & 4       & 6         \\
\rowcolor{gray!10}
Radiation Effects              & 1     & 0        & 0      & 56      & 57        & 1     & 0        & 0      & 23      & 24        \\
Radiotherapy                   & 0     & 0        & 0      & 28      & 28        & 0     & 0        & 0      & 17      & 17        \\
\rowcolor{gray!10}
Rehabilitation                 & 0     & 0        & 1      & 10      & 11        & 0     & 0        & 1      & 10      & 11        \\
Secondary                      & 2     & 0        & 0      & 44      & 46        & 2     & 0        & 0      & 6       & 8         \\
\rowcolor{gray!10}
Standards                      & 2     & 0        & 0      & 67      & 69        & 2     & 0        & 0      & 10      & 12        \\
Statistics \& Numerical Data   & 5     & 1        & 2      & 53      & 61        & 5     & 1        & 2      & 4       & 12        \\
\rowcolor{gray!10}
Supply \& Distribution         & 0     & 0        & 0      & 10      & 10        & 0     & 0        & 0      & 10      & 10        \\
Surgery                        & 5     & 2        & 1      & 749     & 757       & 5     & 2        & 1      & 7       & 15        \\
\rowcolor{gray!10}
Therapeutic Use                & 5     & 0        & 1      & 230     & 236       & 5     & 0        & 1      & 6       & 12        \\
Therapy                        & 45    & 2        & 7      & 373     & 427       & 5     & 2        & 3      & 3       & 13        \\
\rowcolor{gray!10}
Toxicity                       & 2     & 0        & 0      & 29      & 31        & 2     & 0        & 0      & 7       & 9         \\
Transmission                   & 1     & 0        & 1      & 86      & 88        & 1     & 0        & 1      & 15      & 17        \\
\rowcolor{gray!10}
Transplantation                & 0     & 0        & 0      & 31      & 31        & 0     & 0        & 0      & 11      & 11        \\
Trends                         & 0     & 0        & 0      & 1       & 1         & 0     & 0        & 0      & 1       & 1         \\
\rowcolor{gray!10}
Ultrastructure                 & 0     & 0        & 0      & 7       & 7         & 0     & 0        & 0      & 7       & 7         \\
Urine                          & 11    & 0        & 0      & 77      & 88        & 8     & 0        & 0      & 6       & 14        \\
\rowcolor{gray!10}
Veterinary                     & 0     & 0        & 0      & 2       & 2         & 0     & 0        & 0      & 2       & 2         \\
Virology                       & 12    & 0        & 5      & 90      & 107       & 6     & 0        & 5      & 4       & 15       \\
\bottomrule
\end{tabular}%
}
}
\end{table*}

\section{Implementation Details}

\label{app:impl_details}

\subsection{Data Generation Details}

\label{app:impl_details:data_gen}

We generate the reasoning traces and answers using the API of \texttt{deepseek-ai/DeepSeek-R1} model on the \textit{SiliconFlow} platform~\footnote{\url{https://siliconflow.cn/}}. We call the API with its default sampling parameters.
The API calls are scheduled with \texttt{curator}~\footnote{\url{https://github.com/bespokelabsai/curator/}}.

According to the initial observation on the length of the outputs, we set the limit to 8K as no samples have outputs with length larger than 8K.
The prompt is formatted as \verb|"Return your final response within \\boxed{{|
\verb|}}.\n{Question}\n{Options}"|, thus the answers are enclosed and are easy to be extracted and verified. 

We perform data decontamination and dedupliation following OpenThoughts project~\footnote{\url{https://github.com/open-thoughts/open-thoughts/tree/main/open_thoughts}}.

\subsection{SFT Details}

\label{app:impl_details:sft}

All fine-tuning runs use standard language modeling training with `trl` library: we optimize the model to minimize the output cross-entropy on the \verb|reasoning \to answer| sequences (teacher-forcing the entire sequence).
We use the same training hyperparameters as the s1 paper for consistency~\cite{muennighoff2025s1}: 5 epochs of training, a batch size equals 16, a low learning rate of 1e-4 with warmup and cosine decay, a modest weight decay of 1e-4, Adam betas of 0.9 and 0.95. 
The thinking part is enclosed with \verb@<|im_start|>think@ and \verb@<|im_start|>answer@.
SFT is performed with \textit{trl} and \texttt{transformers} libraries.

Training \texttt{m1-7B-1K} and \texttt{m1-7B-23K} is extremely fast on the order of minutes on 8 H100 GPUs, and m1-32B-1K can be trained in a few hours with 16 H100 GPUs. This underscores the efficiency of our approach: unlike massive instruction tuning efforts that require many days on tens of GPU nodes, our models reach convergence with modest compute. We did not apply any reward modeling or RL in training: the model purely learns to imitate the given chain-of-thought format.

\subsection{Evaluation Details}

\label{app:impl_details:eval}

\paragraph{Datasets.}

To thoroughly assess both in-domain performance and generalization, we evaluate on eight medical QA benchmarks, grouped as follows:

In-Distribution Tests: 
\begin{enumerate}
    \item MedMCQA~\cite{pal2022medmcqa} – a collection of ~3.5K multiple-choice questions from Indian medical entrance exams, testing general medical knowledge. 
    \item MedQA-USMLE~\cite{jin2021disease} – the USMLE question dataset (NYU MedQA) containing US medical licensing exam MCQs; we use the standard test split. 
    \item PubMedQA~\cite{jin2019pubmedqa} – a dataset of biomedical research questions (factoid Q paired with abstracts) where the task is to answer yes/no/maybe or short answer. These three were part of our training data pool (though we filtered and sampled from them), so they represent in-domain evaluations. We report accuracy (for MCQ, percentage of correct choices; for PubMedQA, percentage of correct yes/no/maybe).
\end{enumerate}

Out-of-Distribution Tests: 
\begin{enumerate}
    \item MMLU-Pro~\cite{wang2024mmlu} (Medical) – the medical category subset of the Massive Multitask Language Understanding benchmark, which includes professional medicine questions and related subjects. We specifically evaluate on the Professional Medicine section (and report accuracy). We follow the split from~\cite{chen2024huatuogpt}.
    \item GPQA (Medical)~\cite{rein2024gpqa} – the biomedical portion of the Graduate-Level Physics/Chemistry/Biology QA dataset GPQA. This dataset contains extremely challenging, Google-proof multiple-choice questions created by experts, requiring high reasoning (we use the biology/medical questions, ~150 in total). We follow the split from~\cite{chen2024huatuogpt}.
    \item Lancet \& NEJM – we compiled two small sets of QA pairs from The Lancet~\footnote{\url{https://www.thelancet.com/}} and New England Journal of Medicine~\footnote{\url{https://www.nejm.org/}} (NEJM) clinical case reports (answers verified from the text). These assess how models handle medical literature style questions.
    \item MedBullets~\cite{chen2024benchmarking} – a collection of practice questions from the MedBullets medical education platform. We specifically take subsets of difficulty level 4 and 5 (on a 1–5 scale, 5 being hardest), denoted MedBullets Op4 and MedBullets Op5, about 100 questions each, to serve as challenging test sets. 
    \item MedXpertQA~\cite{zuo2025medxpertqa} – a custom set of 50 expert-written multi-step medical reasoning questions we created for qualitative evaluation (free-form answers). For MedXpertQA we report the percentage of questions answered correctly.
    
\end{enumerate}

These out-of-distribution (OOD) sets were not used in training and often differ in style from our training data (e.g. long clinical vignettes, or extremely tricky edge cases). They allow us to test how well \texttt{m1}’s reasoning generalizes.

\paragraph{Methods.}

We compare our models against a broad range of baselines, including both general LLMs and specialized medical LLMs:

\begin{enumerate}
    \item Qwen2.5 Instruct (7B, 32B, 72B) – The base instruct models (no medical fine-tuning). We include these to show the starting performance of the underlying models before our fine-tuning. We also test Qwen2.5 with a chain-of-thought prompting (+CoT), where we simply prompt it to ``think step by step'' at inference, to see if prompting alone can elicit similar reasoning (this baseline uses no additional training).

    \item MedLlama3 (8B)\footnote{\url{https://huggingface.co/johnsnowlabs/}} – An 8B instruction-tuned model released by M42 (Johns Hopkins/APL), one of the early open medical LLMs. We list two versions from their releases.

    \item OpenBioLLM (8B)~\cite{pal2024openbiollms} – The 8B model from Saama AI, fine-tuned with expert-curated medical data.

    \item MMed-Llama (8B)~\cite{qiu2024towards} – A multilingual medical model from MedS3 work, which underwent additional pre-training (denoted MMedS or MMed in results).

    \item Med42 (8B)~\cite{christophe2024med42} – The 8B model from the Med42-v2 suite, instruction and preference-tuned on clinical data.

    \item UltraMedical (8B)~\cite{zhang2024ultramedical} – The 8B model from Tsinghua’s UltraMedical project (we test both the v3.0 and v3.1 versions if available).

    \item HuatuoGPT-o1 (7B \& 8B)~\cite{chen2024huatuogpt} – The smaller versions of HuatuoGPT-o1 (the 70B model’s distilled or intermediate checkpoints) as reported in their paper.

    \item Larger models ($>$10B): We also compare to state-of-the-art open models in the larger size class: Med42-70B~\cite{christophe2024med42}, OpenBioLLM-70B~\cite{pal2024openbiollms}, UltraMedical-70B~\cite{zhang2024ultramedical}, and HuatuoGPT-o1-70B/72B~\cite{chen2024huatuogpt} (if available). These represent the current best open medical LLMs (some claim parity with GPT-4). 

\end{enumerate}

It is worth noting that some of these baselines (e.g. HuatuoGPT-o1~\cite{chen2024huatuogpt}, UltraMedical~
\cite{zhang2024ultramedical}) involve complex training regimes (RL or extensive preference tuning), and in cases like Med42 and OpenBioLLM, they incorporate expert feedback. Our approach does not, so beating or matching them would be a strong indication of the power of test-time scaling.

\paragraph{Inference.}

We use \verb|SGLang| as our inference engine.
We use \verb|bfloat16| precision and greedy sampling (i.e., temperature=0) for inference.
A fixed seed of 42 is used during inference.
The prompt format is: \verb|"{Question}\n{Options}\n{Instruction}"|.
We format options as: \verb|"A. yes\nB. no\nC. maybe"|.
The default instruction is: \verb|"Return your final response within \\boxed| \verb|{{}}."|
For chain-of-thought inference with baseline LLMs Qwen2.5 7B/32B/72B Instruct, we update the instruction to: \verb|"Let's think step by step. Return your final|  \verb|response within \\boxed{{}}."|.

\paragraph{Answer matching.} We try to directly extract the answers from \verb|"\\boxed{{}}"|. If the extraction fails, we follow~\cite{chen2024huatuogpt} to match answers via regex. If multiple answers are matched, we only choose the first one.

\section{Failure Case of Budget Forcing}

\label{app:failure_case_budget_forcing}

\begin{figure}[h]
\begin{center}
\includegraphics[width=\linewidth]{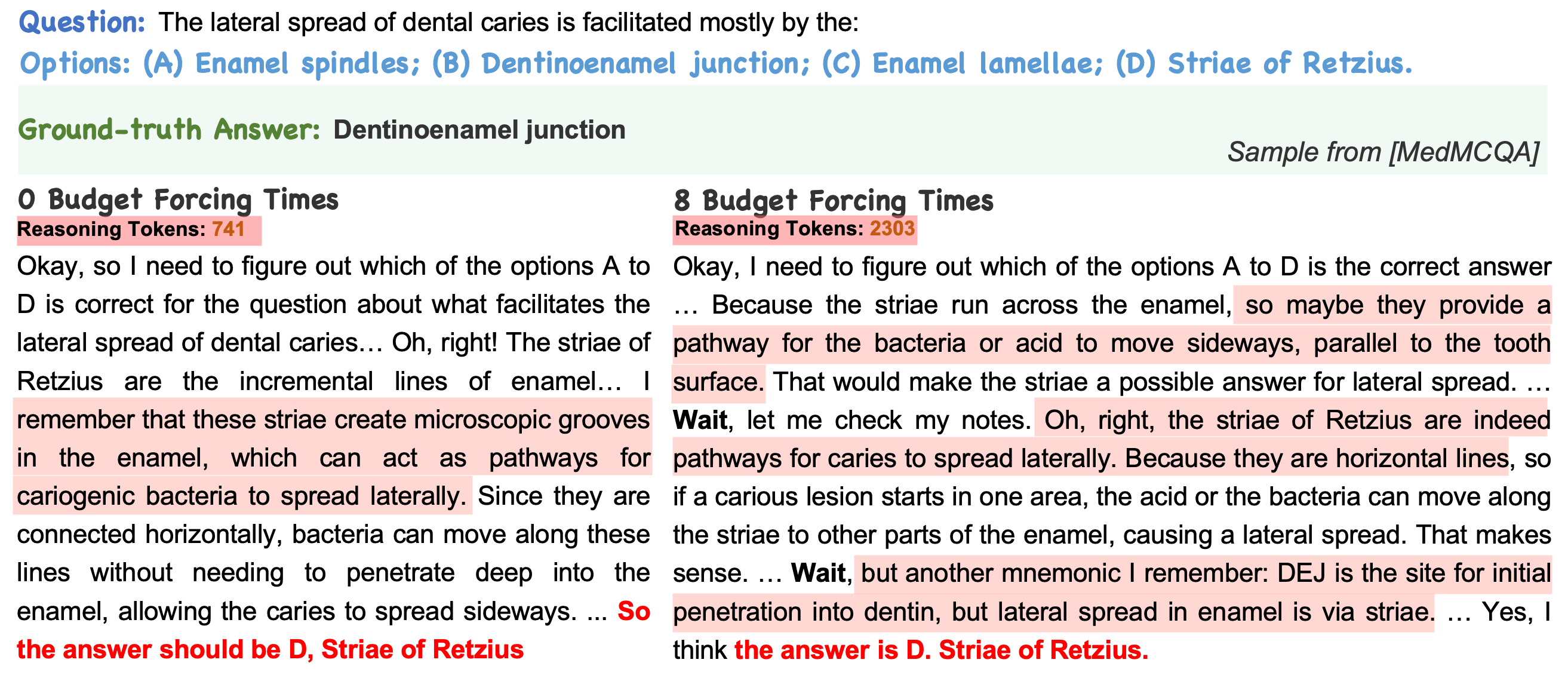}
\end{center}
\caption{
A failure case of budget forcing.
}
\label{fig:case_budget_fail_always}
\end{figure}

We illustrate a failure case of budget forcing in Figure~\ref{fig:case_budget_fail_always}. Initially, the model arrives at the correct answer with concise and accurate reasoning. However, when forced to continue thinking for longer, the extended reasoning introduces confusion and incorporates incorrect anatomical associations, ultimately leading to the wrong answer. This highlights a key limitation of budget forcing in medical QA: more reasoning does not always equate to better reasoning.

\end{document}